\documentclass[journal]{IEEEtran}

\usepackage{mathptmx}
\usepackage{amsmath}
\usepackage{amsthm,amsmath,amssymb}
\usepackage{bm} 
\usepackage{amsfonts}
\usepackage{subeqnarray}
\usepackage{cases,cite}
\usepackage[utf8]{inputenc}
\usepackage{graphicx}
\usepackage{amssymb}
\usepackage{float}
\usepackage{graphicx} 
\usepackage{subfigure}
\usepackage{color}

\IEEEoverridecommandlockouts

\begin{document}

\title{\Large{\bf Game-Theoretic Design of Secure and Resilient Distributed Support Vector Machines with Adversaries}}

\author{Rui~Zhang
        and~Quanyan~Zhu
\thanks{R. Zhang and Q. Zhu are with Department of Electrical and Computer Engineering, New York University, Brooklyn, NY, 11201
E-mail:\{rz885,qz494\}@nyu.edu. 

A preliminary version of this work has been presented at the 18th International Conference on Information fusion, Washington, D.C., 2015 \cite{Zhang}.  } }

\maketitle

\begin{abstract}
With a large number of sensors and control units in networked systems, distributed support vector machines (DSVMs) play a fundamental role in scalable and efficient multi-sensor classification and prediction tasks. However, DSVMs are vulnerable to adversaries who can modify and generate data to deceive the system to misclassification and misprediction. This work aims to design defense strategies for DSVM learner against a potential adversary. We establish a game-theoretic framework to capture the conflicting interests between the DSVM learner and the attacker. The Nash equilibrium of the game allows predicting the outcome of learning algorithms in adversarial environments, and enhancing the resilience of the machine learning through dynamic distributed learning algorithms. We show that the DSVM learner is less vulnerable when he uses a balanced network with fewer nodes and higher degree. We also show that adding more training samples is an efficient defense strategy against an attacker. We present secure and resilient DSVM algorithms with verification method and rejection method, and show their resiliency against adversary with numerical experiments. 
\end{abstract}

\section{Introduction}

Support Vector Machines (SVMs) have been widely used in multi-sensor data fusion problems, such as motor fault detection \cite{banerjee2012multi}, land cover classification \cite{waske2007fusion}, and gas prediction \cite{zhao2009line}. In these applications, a fusion center is required to collect data from each sensor and train the SVM classifier. However, the computations in the fusion center and its communications with sensors become costly when the size of data and network becomes large \cite{durrant1990toward}.

To solve the large-scale data fusion problems, several methods have been developed to speed up SVMs. For example, in \cite{tsang2005core}, Tsang et al. have introduced an approximation method to scale up SVMs. In \cite{dong2005fast}, Dong et al. have presented an efficient SVM algorithm using parallel optimization. These methods only speed up the computations in the fusion center, but the data transmissions between fusion center and sensors still require a significant amount of time and channel usages. 

Efficiency is not the only drawback of the centralized SVM using fusion center. Sensors that collect sensitive or private information to design the classifier may not be willing to share their training data \cite{Forero}. Moreover, a compromised fusion center attacked by an adversary may give erroneous information to all the sensors in the network. Furthermore, compromised sensors may also provide misleading information to the fusion center, and consequently affect uncompromised sensors \cite{chen2008robust}.   

Distributed support vector machines (DSVMs) draw attentions recently as it does not require a fusion center to process data collections and computations \cite{navia2006distributed,Forero,wang2012distributed}. Each node in the network solves decentralized sub-problems themselves using their own data, and only a small amount of data is transferred between nodes, which makes DSVMs more efficient and private than the centralized counterpart. 

However, DSVMs are also vulnerable. For example, misleading information can be spread to the whole network, and the large number of nodes and complex connections in a network makes it harder to detect and track the source of the incorrect information \cite{chan2003security}. Moreover, even though we can find the compromised nodes, an adversary can attack other nodes and spread misleading information. 

Thus, it is important to design secure and resilient distributed support vector machines algorithms against potential attacks from an adversary. In this paper, we focus on a consensus based DSVM algorithm where SVM problem is captured by a set of decentralized convex optimization sub-problems with consensus constraints on their decision variables \cite{Forero,Zhang}. We aim to design defense strategies against potential attacks by analyzing the equilibrium of the game-theoretic model between a DSVM learner and an attacker. 

In our previous work\cite{Zhang}, we have built a game-theoretic framework to capture the conflict of interests between the DSVM learner and the attacker who can modify the training data. In the two-person nonzero-sum game, the learner aims to decentralize the computations over a network of nodes and minimize the error with an effort to minimize misclassification, while the attacker seeks to modify strategically the training data and maximize the error constrained by its computational capabilities. 

The game formulation of the security problem enables
a formal analysis of the impact of the DSVM
algorithm in adversarial environments. The Nash equilibrium
of the game enables the prediction of the outcome, and yields
an optimal response {strategy} to the adversary behaviors. The
game framework also provides a theoretic basis for developing
dynamic learning algorithms that will enhance the security and
the resilience of DSVMs.

In this paper, we propose several defense strategies for a DSVM learner against a potential attacker, and we show the effectiveness of our defense strategies using numerical experiments. The major contributions of this work are multi-fold. 

Firstly, we capture the attacker's objective and constrained capabilities in a game-theoretic framework, and develop
a nonzero-sum game to model the strategic interactions
between an attacker and a learner with a set
of nodes. We then fully characterize the Nash equilibrium by showing
the strategic equivalence between the original nonzerosum
game and a zero-sum game. 

Secondly, we develop secure and resilient distributed algorithms
based on alternating direction method of multipliers
(ADMoM)\cite{Boyd}. Each node communicates with its
neighboring nodes, and updates its decision strategically in response to adversarial environments. We present a summary of numerical results in \cite{Zhang}. 

Lastly, we present four defense strategies against potential attackers. The first defense strategy is to use balanced networks with fewer nodes and higher degrees. In the second defense strategy, we show that adding training samples to compromised nodes can reduce the vulnerability of the learning system. Adding samples to uncompromised nodes at the beginning of the training process also makes the learner less vulnerable. The third defense strategy is to use verification method where each node verifies its received data, and only
accepts reasonable information from neighboring nodes to prevent misleading or illegitimate information sent to uncompromised nodes. The fourth defense strategy is to use rejection method where each node rejects unacceptable updates. Thus, not only misleading information is kept from affecting uncompromised nodes, but also wrong updates could be prevented in compromised nodes. 

{
\subsection{Related Works}
Our work intersects the research areas on data fusion, machine learning, cyber security and machine learning. Machine learning tools have been used to tackle data fusion problems, e.g., \cite{wu2003multi,dalponte2008fusion,hert2006new}. However, machine learning systems can be insecure \cite{barreno2006can}. For example, in \cite{huang2011adversarial}, Huang et al. have shown that SpamBayes and PCA-based network anomaly detection are vulnerable to causative attacks. In \cite{biggio2013evasion}, Biggio et al. have shown that popular classification algorithms can be evaded even if the attacker has limited knowledge of learner's system.}

{
With distributed machine learning tools developed for solving large-scale multi-sensor data fusion problems, each sensor solves sub-problems themselves and transmits information with neighboring sensors \cite{predd2005distributed}. However, cyber security becomes another problem as an attacker may launch malicious cyber attacks to the data fusion networks \cite{chen2009sensor}. Thus, it is important for the machine learning learner to analyze the equilibrium of the game with an adversary and design defense strategies against potential attacks. }

{
Game theory is a natural tool to address this problem. It has been used in the study of the security of machine learning. For example, in \cite{liu2009game}, Liu et al. have modeled the interaction between  a learner and an attacker as a two-person sequential noncooperative Stackelberg game. In \cite{kantarcioglu2008game}, Kantarcioglu et al. have used game theory to analyze the equilibrium behavior of adversarial learning. }

{
Game theory has also been used widely in cyber security as it provides mathematical tools for modeling situations of conflicts and predicting the behaviors of the attacker and defender in network security\cite{manshaei2013game,zhang2017bi}. For example, in \cite{shen2007adaptive}, Shen et al. have built an adaptive Markov game model to infer possible cyber attack patterns. In \cite{jiang2008stochastic}, Jiang et al. have presented an attack prediction and optimal active defense method using a stochastic game. }

{
With game theory, we are able to analyze the game between a distributed machine learning learner with an adversary in a network, and further design defense strategies for the learner against the attacker. In our work, we focus on a class of consensus-based distributed support vector machines algorithms \cite{Forero}. We assume that the attacker has the ability to modify training data to achieve his objectives.  }
 
{
In our previous works \cite{Zhang,zhang2016student,zhang2017game}, we have built a game-theoretic model to capture the conflicts between a DSVM learner and an adversary who can modify training data or labels, and we have solved the game-theoretic problem with ADMoM \cite{Boyd}. In this work, we further analyze the equilibrium behaviors, and design defense strategies for DSVMs against potential attacks. We use numerical experiments to verify the effectiveness of our strategies.  }

\subsection{Organization of the Paper}
The rest of this paper is organized as follows. Section 2 outlines the consensus-based distributed support vector machines.  In Section 3, we establish game-theoretic models for the learner and the attacker. Section 4 deals with the distributed and dynamic algorithms for the learner and the attacker. Section 5 summarizes our previous numerical experiments. Section 6 presents four different defense strategies and their corresponding numerical experiments. Section 7 provides conclusion remarks. 
 
\subsection{Summary of Notations}
Notations in this paper are summarized as follows. Boldface letters are used for matrices (column vectors); $(\cdot)^T$ denotes matrix and vector transposition;
$(\cdot)^{(t)}$ denotes values at step $t$; $[\cdot]_{vu}$ denotes the $vu$-th entry of a matrix; $diag(\mathbf{X})$ is the diagonal matrix with $\mathbf{X}$ on its main diagonal; $\parallel \cdot \parallel$ is the norm of the matrix or vector; $\mathcal{V}$ denotes the set of nodes in a network; $\mathcal{B}_v$ denotes the set of neighboring nodes of node $v$; $\mathcal{U}$ denotes the action set used by the attacker.

\section{Distributed Support Vector Machines}
\begin{figure}
\centering
\subfigure[Network example.]{
\includegraphics[width=0.3\textwidth]{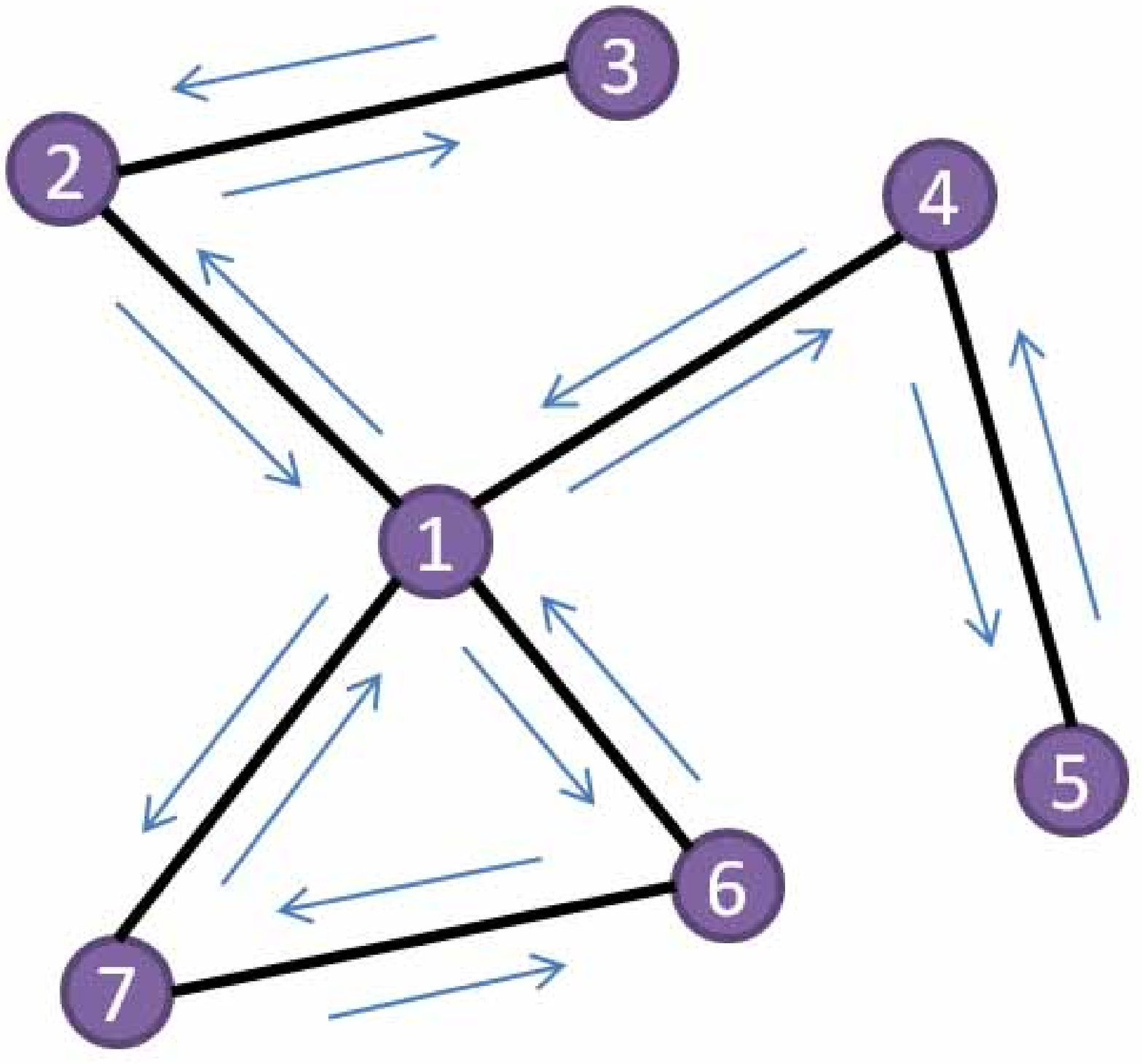}}
\subfigure[SVMs at node $1$.]{
\includegraphics[width=0.4\textwidth]{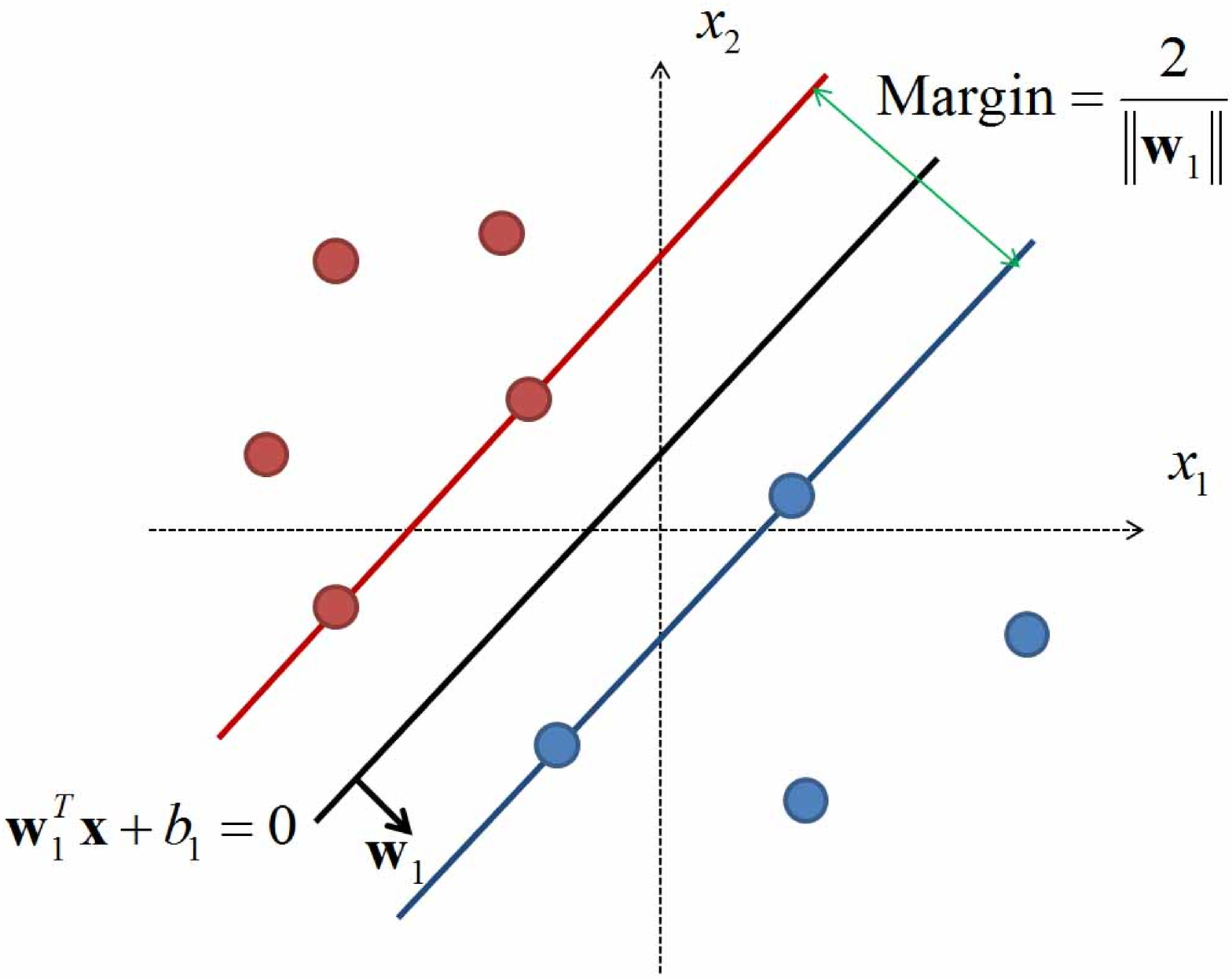}}
\vspace{-2.5mm}
\caption{ Network example. (a) There are $7$ nodes in this network. (b) Each node contains a labelled training set $\mathcal{D}_v:= \lbrace(\mathbf{x}_{vn}, y_{vn}):n=1,...,N_v\rbrace$. Each node can communicate with its neighbors. In each node, the learner aims to find the best linear discriminant line (Black solid line). }
\label{fig:DSVMexample}
\end{figure}

{
In this section, we present a distributed support vector machines learner in the network modeled by an undirected graph $\mathcal{G(V,E)}$ with $\mathcal{V}:=\lbrace 1,...,V \rbrace$ representing the set of nodes, and $\mathcal{E}$ representing the set of links between nodes. Node $v\in \mathcal{V}$ communicates only with his neighboring nodes $\mathcal{B}_v\subseteq\mathcal{V}$. Note that without loss of generality, graph $\mathcal{G}$ is assumed to be connected; in other words, any two nodes in graph $\mathcal{G}$ are connected by a path. However, nodes in $\mathcal{G}$ do not have to be fully connected, which means that nodes are not required to directly connect to all the other nodes in the network. The network can contain cycles. At every node $v\in \mathcal{V}$, a labelled training set $\mathcal{D}_v:= \lbrace(\mathbf{x}_{vn}, y_{vn}):n=1,...,N_v\rbrace$ of size $N_v$ is available, where $\mathbf{x}_{vn} \in \mathbb{R}^p$ represents a
$p$-dimensional pattern, and they are divided into two groups with labels $y_{vn} \in \{+1,-1\}$.} Examples of a network of distributed nodes are illustrated in Fig. \ref{fig:DSVMexample}(a). 

The goal of the learner is to design DSVM algorithms for each node in the network based on its local training data $\mathcal{D}_v$, so that each node has the ability to give new input $\mathbf{x}$ a label of $+1$ or $-1$ without communicating $\mathcal{D}_v$ to other nodes $v^\prime \neq v$. To achieve this, the learner aims to find local maximum-margin linear discriminant functions $ g_v(\mathbf{x}) = \mathbf{x}^T \mathbf{w}_v^{*} + b_v^{*}$ at every node $v\in\mathcal{V}$ with the consensus constraints ${{{\bf{w}}_1^*} = {{\bf{w}}_2^*} =\cdot\cdot\cdot = {{\bf{w}}_{V}^*}},{b_1^*} = {b_2^*} =\cdot\cdot\cdot = {b_{V}^*}$, forcing all the local variables $\{ \mathbf{w}_v^*, b_v^*\}$ to agree across neighboring nodes. Variables $\mathbf{w}_v^*$ and $b_v^*$ of the local discriminant functions $g_v(\mathbf{x})$ can be obtained by solving the following convex optimization problem \cite{Forero}:
\vspace{-2mm}
\begin{equation}
\label{eq:DSVM}
\begin{array}{*{20}{c}}
{\mathop {\min }\limits_{\left\{ {{{\bf{w}}_v},{b_v},\{ {\xi _{vn}}\} } \right\}} \frac{1}{2}\sum\limits_{v \in \mathcal{V}} {{{\left\| {{{\bf{w}}_v}} \right\|}_2^2} + V{C_l}\sum\limits_{v \in \mathcal{V}} {\sum\limits_{n = 1}^{{N_v}} {{\xi _{vn}}} } } }\\

{{\rm{s}}.{\rm{t}}.\begin{array}{*{20}{c}}
{\begin{array}{*{20}{c}}
{{{\rm{y}}_{vn}}({\bf{w}}_v^T{{\bf{x}}_{vn}} + {b_v}) \ge 1 - {\xi _{vn}},}&{\forall v \in {\cal V},n = 1,...,{N_v};}\\
{{\xi _{vn}} \ge 0,}&{\forall v \in {\cal V},n = 1,...,{N_v};}\\
{{{\bf{w}}_v} = {{\bf{w}}_u},{b_v} = {b_u},}&{\forall v \in {\cal V},u \in {\mathcal{B}_v}.}
\end{array}}
\end{array}}
\end{array}
\end{equation}
In the above problem, slack variables $\xi_{vn}$ account for non-linearly separable training sets. $C_l$ is a tunable positive scalar for the learner. 

To solve Problem (\ref{eq:DSVM}), we first define $\mathbf{r}_v:=[{\bf{w}}_v^T,b_v]^T$, the augmented matrix $\mathbf{X}_v:=[(\mathbf{x}_{v1},...,\mathbf{x}_{vN_v})^T,\mathbf{1}_v]$, the diagonal label matrix $\mathbf{Y}_v:=diag([y_{v1},...,y_{vN_v}])$, and the vector of slack variables $\xi_v:=[\xi_{v1},....,\xi_{vN_v}]^T$. With these definitions, it follows readily that ${\bf{w}}_v=(\mathbf{I}_{p+1}-\Pi_{p+1})\mathbf{r}_v$, where $\Pi_{p+1}$ is a $(p+1)\times(p+1)$ matrix with zeros everywhere except for the $(p+1,p+1)$-st entry, given by $[\Pi_{p+1}]_{(p+1)(p+1)}=1$. Thus, Problem (\ref{eq:DSVM}) can be rewritten as
\begin{equation}
\label{eq:DSVMMatrix}
\begin{array}{*{20}{l}}
{\begin{array}{*{20}{l}}
{\mathop {\min }\limits_{\left\{ {{{\bf{r}}_v},{\xi _v},{\omega _{vu}}} \right\}}  \frac{1}{2}\sum\limits_{v \in {\cal V}} {{\bf{r}}_v^T({{\bf{I}}_{p + 1}} - {\Pi _{p + 1}}){{\bf{r}}_v}}  + V{C_l}\sum\limits_{v \in {\cal V}} {{\bf{1}}_v^T{\xi _v}} }\\
\end{array}}\\
{{\rm{s}}.{\rm{t}}.\begin{array}{*{20}{l}}
{\begin{array}{*{20}{l}}
{{{\bf{Y}}_v}{{\bf{X}}_v}{{\bf{r}}_v} \ge {{\bf{1}}_v} - {{\bf{\xi }}_v},}\\
{{{\bf{\xi }}_v} \ge {{\bf{0}}_v},}\\
{{{\bf{r}}_v} = {\omega _{vu}},{\omega _{vu}} = {{\bf{r}}_u},}
\end{array}}&{\begin{array}{*{20}{l}}
{\forall v \in \mathcal{V};}\\
{\forall v \in \mathcal{V};}\\
{\forall v \in \mathcal{V},\forall u \in {\mathcal{B}_v}.}
\end{array}\begin{array}{*{20}{c}}
{}&{\begin{array}{*{20}{c}}
{(\ref{eq:DSVMMatrix}a)}\\
{(\ref{eq:DSVMMatrix}b)}\\
{(\ref{eq:DSVMMatrix}c)}
\end{array}}
\end{array}}
\end{array}}
\end{array}
\end{equation}
Note that $\omega_{vu}$ is used to decompose the decision variable $\mathbf{r}_v$ to its neighbors $\mathbf{r}_u$, where $u \in \mathcal{B}_v$. Problem (\ref{eq:DSVMMatrix}) is a min-problem with matrix form coming from Problem (\ref{eq:DSVM}). 

With alternating direction method of multipliers \cite{Boyd}, Problem (\ref{eq:DSVMMatrix}) can be solved distributed in the following lemma \cite{Forero}, 

\label{Lemma 1}
\noindent
{\bf Lemma 1.}{\it  \ \ With arbitrary initialization $\mathbf{r}_v^{(0)},\lambda_v^{(0)},\omega_{vu}^{(0)}$ and $\alpha_v^{(0)}=\mathbf{0}_{(p+1)\times 1}$, the iterations per node are given by:
\begin{equation}
\label{eq:DSVMSoli1}
\begin{array}{*{20}{l}}
{{\lambda _v^{(t+1)}}}
{ \in \arg \mathop {\max }\limits_{{\bf{0}} \le {{\bf{\lambda }}_v} \le VC_l{{\bf{1}}_v}}  - \frac{1}{2}\lambda _v^T{{\bf{Y}}_v}{{\bf{X}}_v}{\bf{U}}_v^{ - 1}{\bf{X}}_v^T{{\bf{Y}}_v}{\lambda _v}}\\
{\begin{array}{*{20}{c}}
{}&{}
\end{array} \ \ \ \ \ \ \ \ \  \ \ \ \ \ + {{({{\bf{1}}_v} + {{\bf{Y}}_v}{{\bf{X}}_v}{\bf{U}}_v^{ - 1}{{\bf{f}}_v^{(t)}})}^T}{\lambda _v}},
\end{array}
\end{equation}
\begin{equation}
\label{eq:DSVMSoli2}
{{{\bf{r}}_v^{(t+1)}} = {\bf{U}}_v^{ - 1}\left( {{\bf{X}}_v^T{{\bf{Y}}_v}{\lambda _v^{(t+1)}} - {{\bf{f}}_v^{(t)}}} \right)},
\end{equation}
\begin{equation}
\label{eq:DSVMSoli3}
\omega_{vu}^{(t+1)} = \frac{1}{2} (\mathbf{r}_v^{(t+1)} + \mathbf{r}_u^{(t+1)}),
\end{equation}
\begin{equation}
\label{eq:DSVMSoli4}
{{\alpha _v^{(t+1)}} = {\alpha _v^{(t)}} + \frac{\eta }{2}\sum\limits_{u \in {\mathcal{B}_v}} {\left[ {{{\bf{r}}_v^{(t+1)}} - {{\bf{r}}_u^{(t+1)}}} \right]} },
\end{equation}
where $\mathbf{U}_v=(\mathbf{I}_{p+1}-\Pi_{p+1})+2\eta\vert \mathcal{B}_v\vert\mathbf{I}_{p+1},\mathbf{f}_v^{(t)}=2\alpha_v^{(t)}-2\eta\sum_{u\in \mathcal{U}_v}\omega_{vu}^{(t)}$. 
}

\smallskip

The proof of Lemma 1 can be found in \cite{Forero}. Iteration (\ref{eq:DSVMSoli1}) is a quadratic programming problem. $\lambda_v$ are the Lagrange multipliers per node corresponding to constraint (\ref{eq:DSVMMatrix}a). Iteration (\ref{eq:DSVMSoli2}) computes the decision variables $\mathbf{r}_v$, note that the inverse of $\mathbf{U}_v$ always exists and easy to solve. Iteration (\ref{eq:DSVMSoli3}) yields the consensus variables $\omega_{vu}$. Iteration (\ref{eq:DSVMSoli4}) computes $\alpha_v$, e.g., the Lagrange multipliers corresponding to the consensus constraint (\ref{eq:DSVMMatrix}c).  Iterations (\ref{eq:DSVMSoli1})-(\ref{eq:DSVMSoli4}) are summarized into Algorithm 1. Note that at any given iteration $t$ of the algorithm, each node $v\in \mathcal{V}$ computes its own local discriminant function $g_v^{(t)} (\mathbf{x}) $ for any vector $\mathbf{x}$ as 
\begin{equation}
\label{eq:Discriminant}
{g_v^{(t)}({\bf{x}}) = [{{\bf{x}}^T},1]{{\bf{r}}_v^{(t)}}}.
\end{equation}

\begin{table}
\renewcommand{\arraystretch}{1.3}
\label{tablealgorithm1}
\centering
\begin{tabular}{l}
\hline
\bfseries Algorithm 1: ADMoM-DSVM\\
\hline

Randomly initialize $\mathbf{r}_v^{(0)},\lambda_v^{(0)},\omega_{vu}^{(0)}$ 
and $\alpha_v^{(0)}=\mathbf{0}_{(p+1)\times 1}$. \\
1:\ \ \bf{for} $t=0,1,2,...$ do\\
2:\ \ \ \ \ \ \ \ \bf{for all} $v\in \mathcal{V}$ do \\
3:\ \ \ \ \ \ \ \ \ \ \ \ \ Compute $\lambda_v^{(t+1)}$ via (\ref{eq:DSVMSoli1}).\\
4:\ \ \ \ \ \ \ \ \ \ \ \ \ Compute $\mathbf{r}_v^{(t+1)}$ via (\ref{eq:DSVMSoli2}).\\
5:\ \ \ \ \ \ \ \ \bf{end for} \\
6:\ \ \ \ \ \ \ \ \bf{for all} $v\in \mathcal{V}$ do \\
7:\ \ \ \ \ \ \ \ \ \ \ \ Broadcast $\mathbf{r}_v^{(t+1)}$ to all neighbors $u\in \mathcal{B}_v$.\\
8:\ \ \ \ \ \ \ \ \bf{end for}\\
9:\ \ \ \ \ \ \ \ \bf{for all} $v\in \mathcal{V}$ do \\
10:\ \ \ \ \ \ \ \ \ \ \ \ \ Compute $\omega_{vu}^{(t+1)}$ via (\ref{eq:DSVMSoli3}).\\
11:\ \ \ \ \ \ \ \ \ \ \ \ \ Compute $\alpha_v^{(t+1)}$ via (\ref{eq:DSVMSoli4}).\\
12:\ \ \ \ \ \ \ \bf{end for}\\
13:\ \bf{end for}\\
\hline
\end{tabular}
\end{table}

Algorithm 1 solves the DSVM problem using ADMoM technique. It is a fully decentralized network operation, and it does not require exchanging training data or the value of decision functions, which meets the reduced communication overhead and privacy preservation requirements at the same time. However, information transmitted in the network not only helps improve the performance of each node, but also increases the damages from the attacker, as the misleading information can be spread to every node. To design a secure and resilient DSVM algorithm, we first build the attack model to capture the attacker's intentions of breaking the training process of the learner.   

\section{Distributed Support Vector Machines with Adversary} 
{
In this section, we present the game-theoretic framework of a DSVM learner and an attacker who takes over a set of nodes with the aim of breaking the training process of the learner. We assume that the attacker has a complete knowledge of the learner's Problem (\ref{eq:DSVM}) by Kerckhoffs's principle: the enemy knows the system \cite{shannon1949communication}, which enables us to anticipate the interactions of the learner and the attacker in a worst-case scenario. Moreover, the attacker can easily acquire the complete knowledge of the learning systems in reality, for example, by node capture attacks \cite{tague2008modeling} and computer worms \cite{chen2003modeling}, an attacker can compromise the whole network through connections between neighboring nodes, and thus obtain the private and sensitive information of the learner.}

\begin{figure}
\centering
\subfigure[Network under attack.]{
\includegraphics[width=0.3\textwidth]{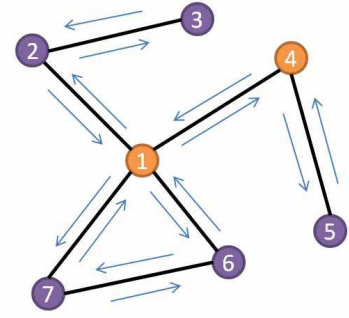}}
\subfigure[SVMs at compromised node $1$.]{
\includegraphics[width=0.4\textwidth]{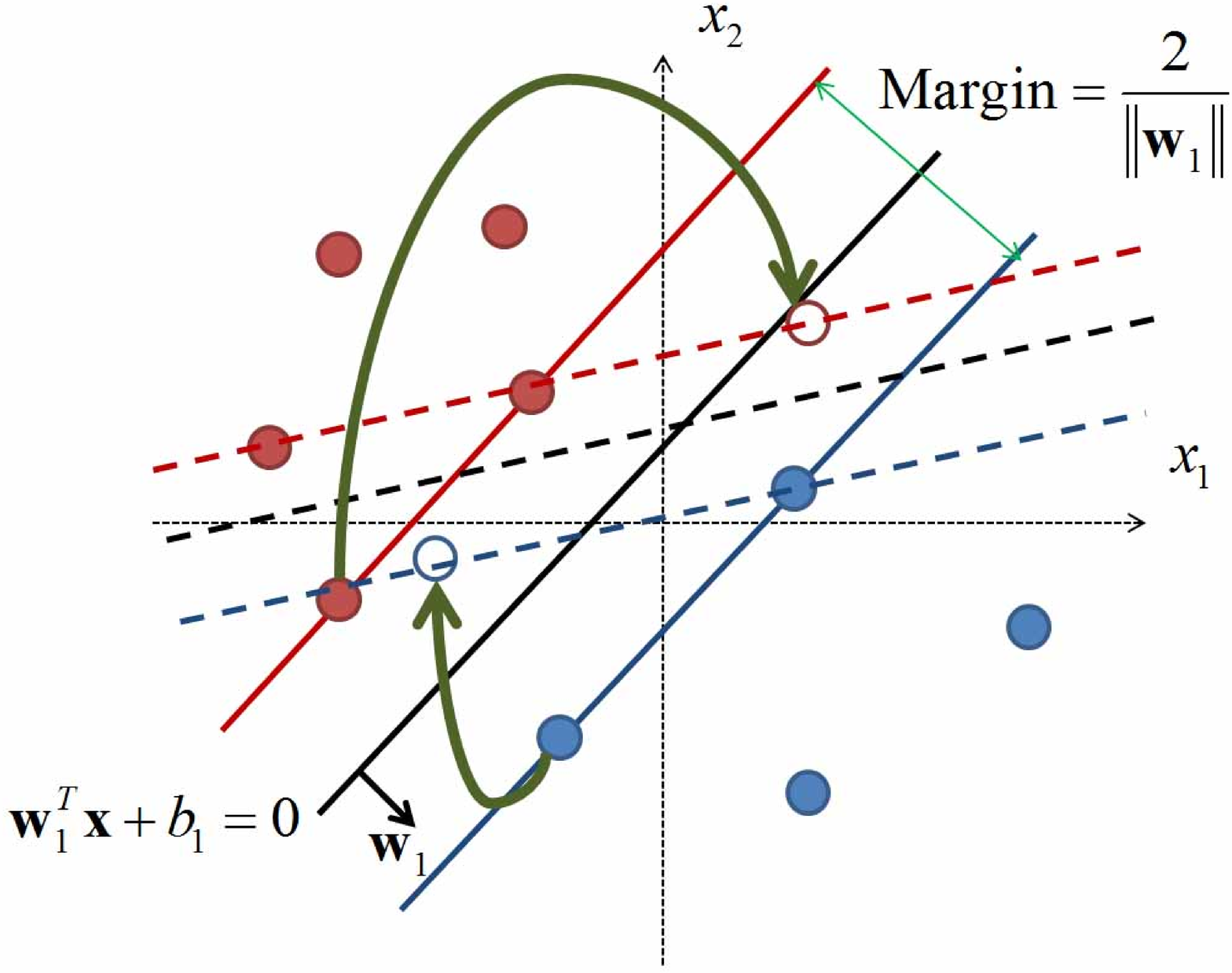}}
\vspace{-2.5mm}
\caption{Network with attacker. (a) Node 1 and 4 are under attack. (b) In compromised node, for example, node 1, an attacker modifies the training data which leads to a wrong linear discriminant line (Black dotted line).}
\label{fig:LAexample}
\end{figure}

{
To achieve the malicious goal, the attacker takes over a set of nodes $\mathcal{V}_a := \{1,...,V_a \}$ and changes $\mathbf{x}_{vn} $ into \[{\widehat {\bf{x}}_{vn}}=\mathbf{x}_{vn} - \delta_{vn},\] where $\delta_{vn} \in \mathcal{U}_v$, and $\mathcal{U}_v$ is the attacker's action set at node $v$. Note that we use $\mathcal{V}_l = \{1,...,V_l\}$ to represent nodes without the attacker. $V = V_a + V_l$ and $\mathcal{V}=\mathcal{V}_l\cup \mathcal{V}_a$. A node in the network is either under attack or not under attack. An example of the impact of the attacker on the learner is shown in Fig. \ref{fig:LAexample}.  This type of attacks represents a challenge for the learner. On the one hand, the learner will find the incorrect classifiers at the compromised nodes, and communications in the network may lead to unanticipated results as misleading information from compromised nodes can be spread to and then used by uncompromised nodes. On the other hand, it is difficult for the learner to identify modified data, and furthermore, in distributed settings, the learner may not even be able to detect which nodes are under attack. Potential real world examples of the attackers are discussed as follows. }

\noindent{
{\bf Example 1}{\it \ \  (Air pollution detection) \cite{khedo2010wireless}.} Consider an air pollution detection system which uses DSVM as the classifiers to determine whether certain areas have air pollution. An attacker can modify the training data of the certain areas to let the system fail to recognize the air pollution. Moreover, the attacker can even modify other areas' training data to achieve his goal, since misleading information can be spread among the whole system by the communications between neighboring nodes. However, with a large amount of training data and areas, the learner will fail to detect the compromised data and areas, and the results of the air pollution detection system will be untrustworthy. }

\noindent{
{\bf Example 2}{\it \ \  (Distributed medical databases) \cite{Forero}.} Suppose several medical centers aim to find classifiers together on a certain disease using DSVM. An attacker can modify the training data of one medical center, which affects not only the compromised medical center, but also the uncompromised medical centers, as the misleading information can be spread among the network. As a result, all the medical centers might give inaccurate diagnosis on the disease. To find out the compromised training data, the learner is required to examine all the training data from all the medical centers, which is costly and sometimes even unrealistic. }

Now Problem (\ref{eq:DSVM}) changes to,
\begin{equation}
\label{eq:aDSVM1}
\begin{array}{l}
\begin{array}{*{20}{c}}
{ \mathop {\min }\limits_{\left\{ {{{\bf{w}}_v},{b_v},\{ \xi_{vn}\}} \right\}}  \frac{1}{2}\sum\limits_{v \in \mathcal{V}} {{{\left\| {{{\bf{w}}_v}} \right\|_2^2}}}  + V{C_l}\sum\limits_{v \in \mathcal{V}} {\sum\limits_{n = 1}^{{N_v}} {{\xi _{vn}}} } }
\end{array}\\
\noindent {\rm{s}}.{\rm{t}}.\\
\begin{array}{*{20}{l}}
{\begin{array}{*{20}{l}}
{{{\rm{y}}_{vn}}({\bf{w}}_v^T{{\bf{x}}_{vn}} + {b_v}) \ge 1 - {\xi _{vn}},}\\{{{\rm{y}}_{vn}}({\bf{w}}_v^T{\widehat{\bf{x}}_{vn}}+ {b_v}) \ge 1 - {\xi _{vn}},}\\
{{\xi _{vn}} \ge 0,}\\
{{{\bf{w}}_v} = {{\bf{w}}_u},{b_v} = {b_u},}
\end{array}}&{\begin{array}{*{20}{l}}
{\forall v \in \mathcal{V}_l,n = 1,...,{N_v};}\\
{\forall v \in \mathcal{V}_a,n = 1,...,{N_v};}\\
{\forall v \in \mathcal{V},n = 1,...,{N_v};}\\
{\forall v \in \mathcal{V},u \in {\mathcal{B}_v}.}
\end{array}}
\end{array}
\end{array}
\end{equation}

By minimizing the objective function in Problem (\ref{eq:aDSVM1}), the learner can obtain the optimal variables $\{\mathbf{w}_v,b_v\}$, which can be used to build up the discriminant function to classify the testing data. The attacker, on the other hand, aims to find an optimal way to modify the data using variables $\{\delta_{vn}\}$ to maximize the same objective function. The behavior of the attacker can thus be captured as follows:
\begin{equation}
\label{eq:aDSVM}
\begin{array}{l}
\begin{array}{*{20}{c}}
{ \mathop {\max }\limits_{\{ {{\delta _{vn}}}\} } \frac{1}{2}\sum\limits_{v \in \mathcal{V}} {{{\left\| {{{\bf{w}}_v}} \right\|_2^2}}}  + V{C_l}\sum\limits_{v \in \mathcal{V}} {\sum\limits_{n = 1}^{{N_v}} {{\xi _{vn}}} } -{C_a}\sum\limits_{v \in \mathcal{V}_a} {\sum\limits_{n = 1}^{{N_v}} {{{\left\| {{\delta _{vn}}} \right\|}_0}} } }
\end{array}\\
\noindent {\rm{s}}.{\rm{t}}.\\
\begin{array}{*{20}{l}}
{\begin{array}{*{20}{l}}
{{{\rm{y}}_{vn}}({\bf{w}}_v^T{{\bf{x}}_{vn}} + {b_v}) \ge 1 - {\xi _{vn}},}\\{{{\rm{y}}_{vn}}({\bf{w}}_v^T({{\bf{x}}_{vn}}-\delta_{vn} )+ {b_v}) \ge 1 - {\xi _{vn}},}\\
{{\xi _{vn}} \ge 0,}\\
{{{\bf{w}}_v} = {{\bf{w}}_u},{b_v} = {b_u},}\\
{{\delta _{vn}} \in {\mathcal{U}_{v}},}
\end{array}}&{\begin{array}{*{20}{l}}
{\forall v \in \mathcal{V}_l,n = 1,...,{N_v};}\\
{\forall v \in \mathcal{V}_a,n = 1,...,{N_v};}\\
{\forall v \in \mathcal{V},n = 1,...,{N_v};}\\
{\forall v \in \mathcal{V},u \in {\mathcal{B}_v};}\\
{\forall v \in {\mathcal{V}_a}.}
\end{array}}
\end{array}
\end{array}
\end{equation}

In above problem, the term $ {C_a}\sum_{v \in \mathcal{V}_a} {\sum_{n = 1}^{{N_v}} {{{\left\| {{\delta _{vn}}} \right\|}_0}} } $ represents the cost function for the attacker. $l_0$ norm is defined as ${\left\| x \right\|_0} := | \{i : {x_i} \neq 0\}|$, i.e., a total number of nonzero elements in a vector. Here, we use the $l_0$ norm to denote the number of elements which are changed by the attacker. The objective function with $l_0$-norm captures the fact that the attacker aims to make the largest impact on the learner by changing the least number of elements. Constraint $\delta_{vn}\in \mathcal{U}_v$ indicates the action set of the attacker. In this paper, we use the following form of $\mathcal{U}_v$: 
\[{\cal U}_v = \left\{ {\left( {{\delta _{v1}},...,{\delta _{vN_v}}} \right)\left| {\sum\limits_{n = 1}^{N_v} {\left\| {{\delta _{vn}}} \right\|_2^2}  \le {C_{v,\delta} }} \right.} \right\},\] 
which is related to the atomic action set $\mathcal{U}_{v0} = \left\{ {\delta_v \left| {\left\| \delta_v  \right\|_2^2 \le {C_{v,\delta} }} \right.} \right\}$. $C_{v,\delta}$ indicates the bound of the sum of the norm of all the changes at node $v$. A higher $C_{v,\delta}$ indicates that the attacker has a large degree of freedom in changing the value $\mathbf{x}_{vn}$. Thus training these data will lead to a higher risk for the learner. Notice that $C_{v,\delta}$ can vary at different nodes, and we use $C_{\delta}$ to represent the situation when $C_{v,\delta}$ are equal at every node. $\delta_v\in\mathbb{R}^p$ from the atomic action set has the same form with $\delta_{vn}$, but $\delta_v$ and $( \delta _{v1},...,\delta _{vN_v} )$ are bounded by same $C_{v,\delta}$. Furthermore, the atomic action set $\mathcal{U}_{v0}$ has the following properties \cite{Xu}.

$\begin{array}{l}
\begin{array}{*{20}{c}}
{({\rm P1  })}&{\bf{0}}
\end{array} \in {\mathcal{U}_{v0}};\\
\begin{array}{*{20}{c}}
{({\rm P2  })}&{{\rm{For \ any \ }}{{\bf{w}}_0} \in {\mathbb{R}^p}:}
\end{array}\\
\begin{array}{*{20}{c}}
{}&{}
\end{array}\mathop {\max }\limits_{\delta_v  \in {\mathcal{U}_{v0}}} \left[ {{\bf{w}}_0^T\delta_v } \right] = \mathop {\max }\limits_{\delta '_v \in {\mathcal{U}_{v0}}} \left[ { - {\bf{w}}_0^T\delta'_v} \right] <  + \infty .
\end{array}$

The first property (P1) states that the attacker cannot choose to change the value of $\mathbf{x}_{vn}$. Property (P2) states that the atomic action set is bounded and symmetric. Here, ``bounded'' means that the attacker has the limit on the capability of changing $\mathbf{x}_{vn}$. It is reasonable since changing the value significantly will result in the evident detection of the learner. 

For the learner, the learning process is to find the discriminant function which separates the training data into two classes with less error, and then use the discriminant function to classify testing data. Since the attacker has the ability to change the value of original data $\mathbf{x}_{vn}\in \mathcal{X}$ into ${\widehat {\bf{x}}_{vn}} \in \widehat{\mathcal{X}}$, the learner will find the discriminant function that separates the data in $\widehat{\mathcal{X}}$ more accurate, rather than the data in $\mathcal{X}$. As a result, when using the discriminant function to classify the testing data $\mathbf{x}\in\mathcal{X}$, it will be prone to be misclassified. 

{
Since the learner aims at a high classification accuracy, while the attacker seeks to lower the accuracy, we can capture the conflicting goals of the players as a two-person nonzero-sum game by combining Problem (\ref{eq:aDSVM1}) and Problem (\ref{eq:aDSVM}) together. The solution to the game problem is described by Nash equilibrium, which yields the equilibrium strategies for both players, and predicts the outcome of machine learning in the adversarial environment. By comparing Problem (\ref{eq:aDSVM1}) with Problem (\ref{eq:aDSVM}), we notice that they contain the same terms in their objective functions and the constraints in the two problems are uncoupled. As a result, the nonzero-sum game can be reformulated into a zero-sum game, which takes the minimax or max-min form as follows: 
\begin{equation}
\label{eq:alminmax}
\begin{array}{l}
\begin{array}{*{20}{l}}
{\mathop {\min }\limits_{\left\{ {{{\bf{w}}_v},{b_v},\{\xi_{vn}\}} \right\}} \mathop {\max }\limits_{\{ {\delta _{vn}}\} } \frac{1}{2}\sum\limits_{v \in \mathcal{V}} {{{\left\| {{{\bf{w}}_v}} \right\|_2^2}}}  + V{C_l}\sum\limits_{v \in \mathcal{V}} {\sum\limits_{n = 1}^{{N_v}} {{\xi _{vn}}} } }\\
{\begin{array}{*{20}{c}}
{}&{}
\end{array} \ \ \ \ \ \ \ \ \ \ \ \ \ \ -{C_a}\sum\limits_{v \in \mathcal{V}_a} {\sum\limits_{n = 1}^{{N_v}} {{{\left\| {{\delta _{vn}}} \right\|}_0}} } }
\end{array}\\
{\rm{s}}.{\rm{t}}.\\
\begin{array}{*{20}{l}}
{\begin{array}{*{20}{l}}
{{{\rm{y}}_{vn}}({\bf{w}}_v^T{{\bf{x}}_{vn}} + {b_v}) \ge 1 - {\xi _{vn}},}\\{{{\rm{y}}_{vn}}({\bf{w}}_v^T({{\bf{x}}_{vn}}-\delta_{vn} )+ {b_v}) \ge 1 - {\xi _{vn}},}\\
{{\xi _{vn}} \ge 0,}\\
{{{\bf{w}}_v} = {{\bf{w}}_u},{b_v} = {b_u},}\\
{{\delta _{vn}} \in {\mathcal{U}_{v}},}
\end{array}}&{\begin{array}{*{20}{l}}
{\forall v \in \mathcal{V}_l,n = 1,...,{N_v};}\\
{\forall v \in \mathcal{V}_a,n = 1,...,{N_v};}\\
{\forall v \in \mathcal{V},n = 1,...,{N_v};}\\
{\forall v \in \mathcal{V},u \in {\mathcal{B}_v};}\\
{\forall v \in {\mathcal{V}_a}.}
\end{array}}
\end{array}
\end{array}
\end{equation} 
Note that the first and fourth constraints only contribute to the minimization part of the problem, the fifth constraint only affects the maximization part. The second and third constraints contribute to both the minimization and the maximization part. The first term of the objective function is the inverse of the distance of margin. The second term is the sum of all the slack variables which captures the error penalties. On one hand, minimizing the objective function captures the trade-off between a larger margin and a small error penalty of the learner, while on the other hand, maximizing the objective function captures the trade-off between a large error penalty and a small cost of the attacker. As a result, solving Problem (\ref{eq:alminmax}) can be understood as finding the saddle-point equilibrium of the zero-sum game between the attacker and the learner.}

\label{definitino1}
\noindent{ {\bf Definition 1.}{\it \ \ Let $\mathcal{S}_L $ and $ \mathcal{S}_A$ be the action sets for the DSVM learner and the attacker, respectively. Notice that here $\mathcal{S}_A = \{ \mathcal{U}_v \}_{v\in \mathcal{V}_a}$. Then, the strategy pair $\left( {\left\{ {{{\bf{w}}_v^*},{b_v^*},\{\xi_{vn}^*  \}} \right\},\left\{ {{\delta _{vn}^*}} \right\}} \right)$ is a saddle-point equilibrium solution of the zero-sum game defined by the triple ${G_z} := \left\langle {\left\{ {L,A} \right\},\left\{ {{\mathcal{S}_L},{\mathcal{S}_A}} \right\},K} \right\rangle $, if
$K\left( {\left\{ {{{\bf{w}}_v^*},{b_v^*},\{\xi_{vn}^*  \}} \right\},\left\{ {{\delta _{vn}}} \right\}} \right) \leq   K\left( {\left\{ {{{\bf{w}}_v^*},{b_v^*},\{\xi_{vn}^*  \}} \right\},\left\{ {{\delta _{vn}^*}} \right\}} \right) \leq K\left( {\left\{ {{{\bf{w}}_v},{b_v},\{\xi_{vn}  \}} \right\},\left\{ {{\delta _{vn}^*}} \right\}} \right),\forall v \in \mathcal{V} $,
where $K$ is the objective function of Problem (10).
} }

Based on the property of the action set and atomic action set, Problem (\ref{eq:alminmax}) can be further simplified as stated in the following lemma \cite{Zhang}.

\label{Lemma 2}
\noindent
{\bf Lemma 2.}{\it \ \ Assume that $\mathcal{U}_v$ is an action set with corresponding atomic action set $\mathcal{U}_{v0}$. Then, Problem (\ref{eq:alminmax}) is equivalent to the following optimization problem:
\begin{equation}
\label{eq:MinMax}
\begin{array}{l}
\begin{array}{*{20}{l}}
{\mathop {\min }\limits_{\left\{ {{{\bf{w}}_v},{b_v},\{\xi_{vn}\}} \right\}} \mathop {\max }\limits_{\{ {\delta _v}\} } \frac{1}{2}\sum\limits_{v \in \mathcal{V}} {{{\left\| {{{\bf{w}}_v}} \right\|_2^2}}}  + V{C_l}\sum\limits_{v \in \mathcal{V}} {\sum\limits_{n = 1}^{{N_v}} {{\xi _{vn}}} } }\\
{\begin{array}{*{20}{c}}
{}&{}
\end{array} \ \ \ \ \ \ \ \ \ \ \ + \sum\limits_{v \in \mathcal{V}_a} {\left( {{V_a}{C_l}{\bf{w}}_v^T{\delta _v} - {C_a}{{\left\| {{\delta _v}} \right\|}_0}} \right)} }
\end{array}\\
{\rm{s}}.{\rm{t}}.\\
\begin{array}{*{20}{c}}
{\begin{array}{*{20}{l}}
{{{\rm{y}}_{vn}}({\bf{w}}_v^T{{\bf{x}}_{vn}} + {b_v}) \ge 1 - {\xi _{vn}},}\\
{{\xi _{vn}} \ge 0,}\\
{{{\bf{w}}_v} = {{\bf{w}}_u},{b_v} = {b_u},}\\
{{\delta _v} \in {\mathcal{U}_{v0}},}
\end{array}}&{\begin{array}{*{20}{l}}
{\forall v \in \mathcal{V},n = 1,...,{N_v};}\\
{\forall v \in \mathcal{V},n = 1,...,{N_v};}\\
{\forall v \in \mathcal{V},u \in {\mathcal{B}_v};}\\
{\forall v \in {\mathcal{V}_a}.}
\end{array}}
\end{array}
\end{array}
\end{equation}}
{
\begin{proof}
See Appendix A. 
\end{proof}}
 
In Problem (\ref{eq:alminmax}), the second and third constraints are the coupled terms with the second term of the objective function. But in Problem (\ref{eq:MinMax}), the only coupled term is $V_a C_l \mathbf{w}_v^T \delta_v$, which is linear in the decision variables of the attacker and the learner, respectively.

\section{ADMoM-DSVM and Distributed Algorithm}

In the previous section, we have combined Problem (\ref{eq:aDSVM1}) for the learner with Problem (\ref{eq:aDSVM}) for the attacker into one minimax Problem (\ref{eq:alminmax}), and have showed its equivalence to Problem (\ref{eq:MinMax}). In this section, we develop iterative algorithms to find equilibrium solutions to Problem (\ref{eq:MinMax}). Using a similar method in Section II, Problem (\ref{eq:MinMax}) can be rewritten into matrix form as
\begin{equation}
\label{eq:MinMaxMatrix}
\begin{array}{*{20}{l}}
{\begin{array}{*{20}{l}}
{\mathop {\min }\limits_{\left\{ {{{\bf{r}}_v},{\xi _v},{\omega _{vu}}} \right\}} \mathop {\max }\limits_{\{ {\delta _v}\} } \frac{1}{2}\sum\limits_{v \in {\cal V}} {{\bf{r}}_v^T({{\bf{I}}_{p + 1}} - {\Pi _{p + 1}}){{\bf{r}}_v}}  + V{C_l}\sum\limits_{v \in {\cal V}} {{\bf{1}}_v^T{\xi _v}} }\\
{\begin{array}{*{20}{c}}
{}&{\begin{array}{*{20}{c}}
{}&{}
\end{array}}
\end{array} + \sum\limits_{v \in {{\cal V}_a}} {\left( {{V_a}{C_l}{\bf{r}}_v^T({{\bf{I}}_{p + 1}} - {\Pi _{p + 1}}){\delta _v} - {C_a}{{\left\| {{\delta _v}} \right\|}_0}} \right)} }
\end{array}}\\
{{\rm{s}}.{\rm{t}}.\begin{array}{*{20}{l}}
{\begin{array}{*{20}{l}}
{{{\bf{Y}}_v}{{\bf{X}}_v}{{\bf{r}}_v} \ge {{\bf{1}}_v} - {{\bf{\xi }}_v},}\\
{{{\bf{\xi }}_v} \ge {{\bf{0}}_v},}\\
{{{\bf{r}}_v} = {\omega _{vu}},{\omega _{vu}} = {{\bf{r}}_u},}\\
{{\delta _v} \in {{\cal U}_{v0}},}
\end{array}}&{\begin{array}{*{20}{l}}
{\forall v \in \mathcal{V};}\\
{\forall v \in \mathcal{V};}\\
{\forall v \in \mathcal{V},\forall u \in {\mathcal{B}_v};}\\
{\forall v \in {\mathcal{V}_a}.}
\end{array}\begin{array}{*{20}{c}}
{}&{\begin{array}{*{20}{c}}
{(\ref{eq:MinMaxMatrix}a)}\\
{(\ref{eq:MinMaxMatrix}b)}\\
{(\ref{eq:MinMaxMatrix}c)}\\
{(\ref{eq:MinMaxMatrix}d)}
\end{array}}
\end{array}}
\end{array}}
\end{array}
\end{equation} 

To solve problem (\ref{eq:MinMaxMatrix}), we use best response dynamics to construct the best response for the min-problem and max-problem separately. The min-problem and max-problem are archived by fixing $\{ \delta_v \}$ and $\{ \mathbf{r}_v \}$, respectively. With ADMoM \cite{Eckstein}, we can develop a method of solving Problem (\ref{eq:MinMaxMatrix}) in a distributed way as follows: The first step is that each node randomly picks an initial $ \mathbf{r}_v^{(0)}, \delta_{v}^{(0)} $ and $ \alpha_v = \mathbf{0}_{(p+1)\times 1} $, then solve the max-problem with $\{ \mathbf{r}_v^{(0)} \}$, and obtain$\{\delta_{v}^{(1)}\}$. The next step is to solve the min-problem with $\{\delta_{v}^{(1)}\}$ and obtain $\{ \mathbf{r}_v^{(1)} \}$, then we repeat solving the max-problem with $\{ \mathbf{r}_v \}$ from the previous step and the min-problem with $\{\delta_{v} \}$ from the previous step until the pair $\{ \mathbf{r}_v, \delta_{v} \}$ achieves convergence. The iterations of solving Problem (\ref{eq:MinMaxMatrix}) can be summarized as follows{\cite{Zhang}}.

\label{Lemma 3}
\noindent
{\bf Lemma 3.}{\it  \ \ With arbitrary initialization $\mathbf{\delta}_v^{(0)},\mathbf{r}_v^{(0)},\lambda_v^{(0)},\omega_{vu}^{(0)}$ and $\alpha_v^{(0)}=\mathbf{0}_{(p+1)\times 1}$, the iterations per node are given by:
\begin{equation}
\label{eq:MinMaxSoli1}
\begin{array}{l}
{\delta _v^{(t+1)}} \in \arg \mathop {\max }\limits_{\left\{ {{\delta _v},{s_v}} \right\}} {V_a}{C_l}{\bf{r}}_v^{(t)T}({{\bf{I}}_{p + 1}} - {\Pi _{p + 1}}){\delta _v}\\
\begin{array}{*{20}{c}}
{\begin{array}{*{20}{c}}
{\begin{array}{*{20}{c}}
{}&{}
\end{array}}&{}
\end{array}}&{}
\end{array} \ \ \ \ \ \ \ \ \ \  - {{\bf{1}}^T}{s_v}\\
{\rm{s}}.{\rm{t}}.{\rm{   }}\begin{array}{*{20}{c}}
{\begin{array}{*{20}{l}}
{{C_a}{\delta _v} \le {s_v},}\\
{{C_a}{\delta _v} \ge  - {s_v},}\\
{{\delta _v} \in {{\cal U}_{v0}},}
\end{array}}&{\begin{array}{*{20}{l}}
{\forall v \in {\cal V}_a;}\\
{\forall v \in {\cal V}_a;}\\
{\forall v \in {\cal V}_a.}
\end{array}}
\end{array}
\end{array}
\end{equation}
\begin{equation}
\label{eq:MinMaxSoli2}
\begin{array}{*{20}{l}}
{{\lambda _v^{(t+1)}}}
{ \in \arg \mathop {\max }\limits_{{\bf{0}} \le {{\bf{\lambda }}_v} \le VC_l{{\bf{1}}_v}}  - \frac{1}{2}\lambda _v^T{{\bf{Y}}_v}{{\bf{X}}_v}{\bf{U}}_v^{ - 1}{\bf{X}}_v^T{{\bf{Y}}_v}{\lambda _v}}\\
{\begin{array}{*{20}{c}}
{}&{}
\end{array} \ \ \ \ \ \ \ \ \  \ \ \ \ \ + {{({{\bf{1}}_v} + {{\bf{Y}}_v}{{\bf{X}}_v}{\bf{U}}_v^{ - 1}{{\bf{f}}_v^{(t)}})}^T}{\lambda _v}},
\end{array}
\end{equation}
\begin{equation}
\label{eq:MinMaxSoli3}
{{{\bf{r}}_v^{(t+1)}} = {\bf{U}}_v^{ - 1}\left( {{\bf{X}}_v^T{{\bf{Y}}_v}{\lambda _v^{(t+1)}} - {{\bf{f}}_v^{(t)}}} \right)},
\end{equation}
\begin{equation}
\label{eq:MinMaxSoli4}
\omega_{vu}^{(t+1)} = \frac{1}{2} (\mathbf{r}_v^{(t+1)} + \mathbf{r}_u^{(t+1)}),
\end{equation}
\begin{equation}
\label{eq:MinMaxSoli5}
{{\alpha _v^{(t+1)}} = {\alpha _v^{(t)}} + \frac{\eta }{2}\sum\limits_{u \in {\mathcal{B}_v}} {\left[ {{{\bf{r}}_v^{(t+1)}} - {{\bf{r}}_u^{(t+1)}}} \right]} },
\end{equation}
where $\mathbf{U}_v=(\mathbf{I}_{p+1}-\Pi_{p+1})+2\eta\vert \mathcal{B}_v\vert\mathbf{I}_{p+1},\mathbf{f}_v^{(t)}=V_a C_l\delta_v^{(t)}+2\alpha_v^{(t)}-2\eta\sum_{u\in \mathcal{U}_v}\omega_{vu}^{(t)}$. 
}
{
\begin{proof}
See Appendix B. 
\end{proof}}

Iterations (\ref{eq:MinMaxSoli1}) corresponds to the attacker's Max-Problem (\ref{eq:aDSVM}), while iterations (\ref{eq:MinMaxSoli2})-(\ref{eq:MinMaxSoli5}) corresponds to the learner's Min-Problem (\ref{eq:aDSVM1}). The Minimax Problem (\ref{eq:MinMax}) is solved  by iterating them together. Note that, iterations (\ref{eq:MinMaxSoli2})-(\ref{eq:MinMaxSoli5}) differ from iterations (\ref{eq:DSVMSoli1})-(\ref{eq:DSVMSoli4}) only in $\mathbf{f}_v$. In (\ref{eq:MinMaxSoli2})-(\ref{eq:MinMaxSoli5}), $\mathbf{f}_v$ adds another term $V_a C_l \delta_v$ which captures the attacker's impact on the learner. Iterations (\ref{eq:MinMaxSoli1})-(\ref{eq:MinMaxSoli5}) are summarized into Algorithm 2.

\begin{table}[H]
\renewcommand{\arraystretch}{1.3}
\label{tablealgorithm2}
\centering
\begin{tabular}{l}
\hline
\bfseries Algorithm 2: DSVM under attack\\
\hline

Randomly initialize $\mathbf{\delta}_v^{(0)},\mathbf{r}_v^{(0)},\lambda_v^{(0)},\omega_{vu}^{(0)}$ 
and $\alpha_v^{(0)}=\mathbf{0}_{(p+1)\times 1}$. \\
1:\ \ \bf{for} $t=0,1,2,...$ do\\
2:\ \ \ \ \ \ \ \ \bf{for all} $v\in \mathcal{V}$ do \\
3:\ \ \ \ \ \ \ \ \ \ \ \ \ Compute $\delta_v^{(t+1)}$ via (\ref{eq:MinMaxSoli1}).\\
4:\ \ \ \ \ \ \ \ \bf{end for} \\
5:\ \ \ \ \ \ \ \ \bf{for all} $v\in \mathcal{V}$ do \\
6:\ \ \ \ \ \ \ \ \ \ \ \ \ Compute $\lambda_v^{(t+1)}$ via (\ref{eq:MinMaxSoli2}).\\
7:\ \ \ \ \ \ \ \ \ \ \ \ \ Compute $\mathbf{r}_v^{(t+1)}$ via (\ref{eq:MinMaxSoli3}).\\
8:\ \ \ \ \ \ \ \ \bf{end for} \\
9:\ \ \ \ \ \ \ \ \bf{for all} $v\in \mathcal{V}$ do \\
10:\ \ \ \ \ \ \ \ \ \ \ \ Broadcast $\mathbf{r}_v^{(t+1)}$ to all neighbors $u\in \mathcal{B}_v$.\\
11:\ \ \ \ \ \ \ \bf{end for}\\
12:\ \ \ \ \ \ \ \bf{for all} $v\in \mathcal{V}$ do \\
13:\ \ \ \ \ \ \ \ \ \ \ \ \ Compute $\omega_{vu}^{(t+1)}$ via (\ref{eq:MinMaxSoli4}).\\
14:\ \ \ \ \ \ \ \ \ \ \ \ \ Compute $\alpha_v^{(t+1)}$ via (\ref{eq:MinMaxSoli5}).\\
15:\ \ \ \ \ \ \ \bf{end for}\\
16:\ \bf{end for}\\
\hline
\end{tabular}
\end{table}

Algorithm 2 solves the Minimax Problem (\ref{eq:MinMax}) using ADMoM technique. It is a fully distributed algorithm which only requires transmitting $\mathbf{r}_v$ between each nodes. The attacker's behavior is captured as calculating $\delta_v$ by solving the linear programming Problem (\ref{eq:MinMaxSoli1}) with the learner's decision variable $\mathbf{r}_v$. The learner's behavior is captured as computing (\ref{eq:MinMaxSoli2})-(\ref{eq:MinMaxSoli5}) with $\delta_v$ from the attacker. Since the learner transmits $\mathbf{r}_v$ to each neighboring nodes, misleading information will eventually spread in the whole network, which leads to misclassifications in all nodes. 

\section{Numerical Results}

In this section, we summarize numerical results of DSVM under adversarial environments. We use empirical risk to measure the performance of DSVM. The empirical risk at node $v$ at step $t$ is defined as follows:
\begin{equation}
\label{eq:ErrorInNodev}
{{\bf{R}}_v^{(t)}}: = \frac{1}{{{\widetilde N_{v}}}}\sum\limits_{n = 1}^{{\widetilde N_{v}}} {\frac{1}{2}\left| {{{\widetilde y}_{vn}} - {{\widehat y}_{vn}^{(t)}}} \right|} ,
\end{equation} 
where ${{{\widetilde y}_{vn}}}$ is the true label; ${{{\widehat y}_{vn}^{(t)}}}$ is the predicted label; and $\widetilde N_v$ represents the number of testing samples in node $v$. The empirical risk (\ref{eq:ErrorInNodev}) sums over the number of misclassified samples in node $v$, and then divides it by the number of all testing samples in node $v$. Notice that testing samples can vary for different nodes. In order to measure the global performance, we use the global empirical risk defined as follows:   
\begin{equation}
\label{eq:Error}
{{\bf{R}}_{G}^{(t)}}: = \frac{1}{{\widetilde N}}\sum\limits_{v \in \mathcal{V}} {\sum\limits_{n = 1}^{{{\widetilde N}_v}} {\frac{1}{2}\left| {{{\widetilde y}_{vn}} - {{\widehat y}_{vn}^{(t)}}} \right|} } ,
\end{equation} 
where ${\widetilde N} = \sum\limits_{v \in \mathcal{V}} {\widetilde N_v}$, representing the total number of testing samples. Clearly, a higher global empirical risk shows that there are more testing samples being misclassified, i.e., a worse performance of DSVM. We use the first experiment to illustrate the significant impact of the attacker.

\begin{figure}
\centering
\subfigure[]{
\includegraphics[width=0.4\textwidth]{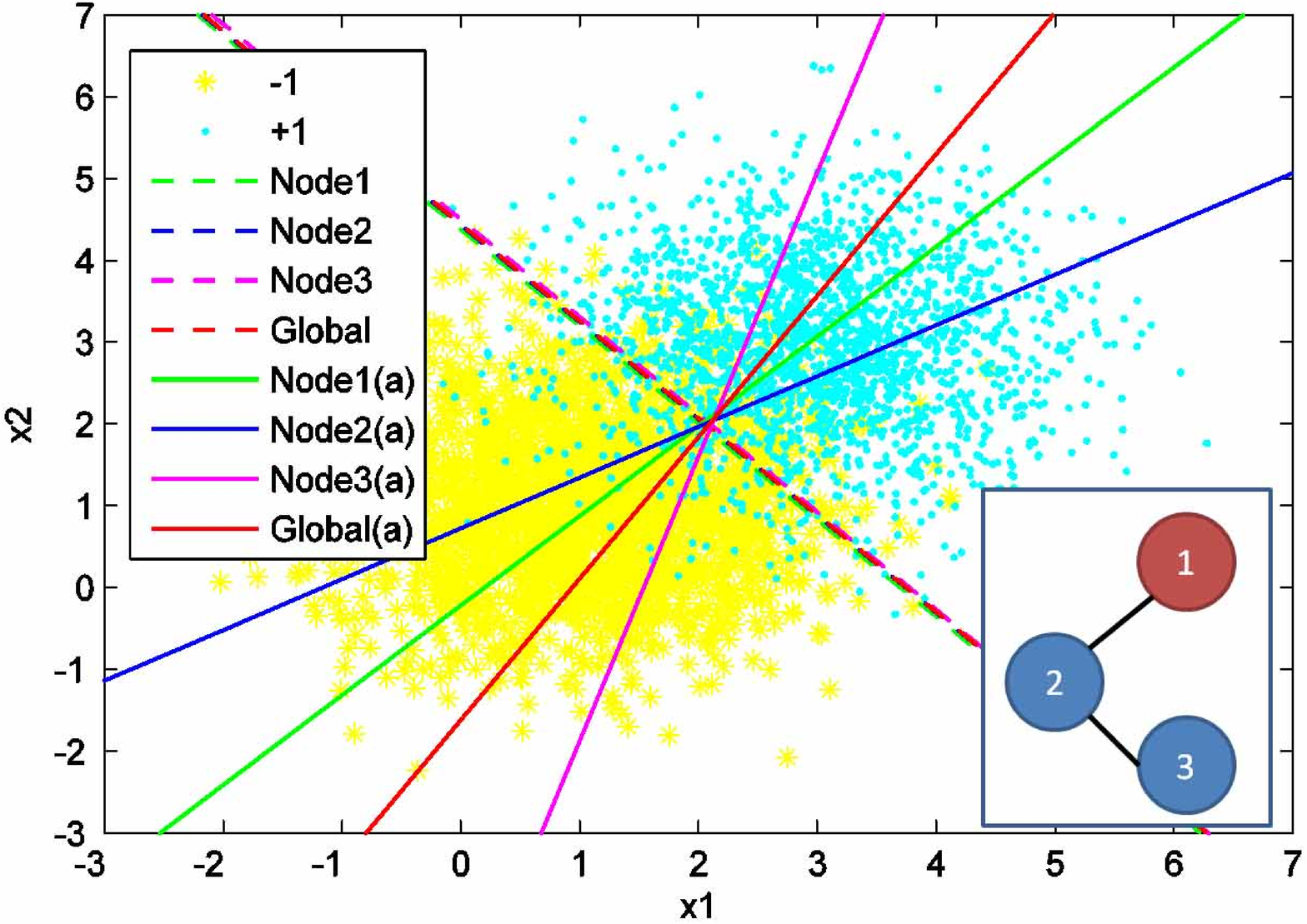}}
\subfigure[]{
\includegraphics[width=0.4\textwidth]{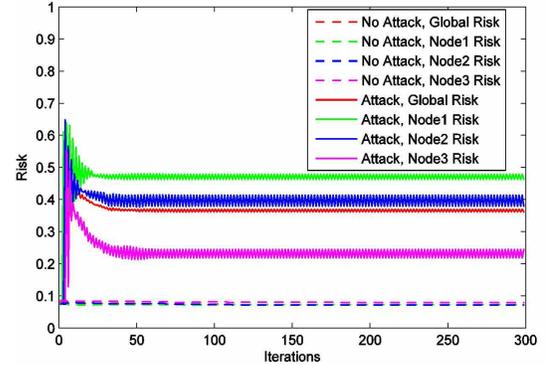}}
\vspace{-3mm}
\caption{Evolution of the empirical risks of ADMoM-DSVM with the attacker at a network with $3$ nodes shown at the bottom right corner of figure (a). An attacker only attacks red node $1$ from the beginning of the training process. Training data and testing data are generated from two Gaussian classes. Dotted lines and solid lines show the results when there is no attacker and there is an attacker, respectively. Different colors represent risks or discriminant lines of different nodes.}
\label{fig:NumericalExample}
\end{figure}

Consider a network with $3$ nodes, which can be seen at the bottom right corner of Fig. \ref{fig:NumericalExample}(a). Each node contains $80$ training samples and $1000$ testing samples from the same global training dataset, which is shown as points and stars in Fig. \ref{fig:NumericalExample}(a). Yellow stars and magenta points are labelled as $-1$ and $+1$, respectively. They are generated from two-dimensional Gaussian distributions with mean vectors $[1,1]$ and $[3,3]$, with the same covariance matrix $[1,0;0,1]$. The learner has the ability $C_l = 1$ and $\eta = 1$. The attacker has the atomic action set parameter $C_{1,\delta}
 = 9,000,000 $, and the cost parameter $C_a = 1$. The attacker only attacks Node 1 which is shown as red points at the bottom right corner in Fig. \ref{fig:NumericalExample}(a) and attack starts from the beginning of the training process. Numerical results are shown in Fig. \ref{fig:NumericalExample}(b). Notice that the risks when there is an attacker are much higher than the risks when there is no attacker, which indicates that the attacker has a significant impact on the learner. Also, we can conclude that the risks at the node under attack are much higher than the risks in nodes without attack, but both of them are higher than the risks when there is no attacker in the network. This shows that the attacker has the ability to affect uncompromised nodes through network connections. We can also observe from Fig. \ref{fig:NumericalExample}(a) that the solid lines, which represent the situation when there is an attacker, cannot separate yellow {stars} and magenta points.

It is clear that the attacker can cause disastrous results for the learner. In our previous work \cite{Zhang}, we have shown that results of the game between the DSVM learner and the attacker are affected by both the attacker's ability and the network topologies. We summarize our previous numerical results from \cite{Zhang} in the following observations. 

\label{Observation1}
\noindent
{\bf Observation 1.}{\it \ \ \ The attacker's ability is captured by four measures, i.e., $(i)$ the time $t$ for the attacker to take an action, $(ii)$ the atomic action set parameter $C_{v,\delta}$, $(iii)$ the cost parameter $C_a$, and $(iv)$ the number of compromised nodes $| \mathcal{V}_a |$. The impact of them is summarized as follows.
\begin{itemize}
\item The time $t$ for an attacker to take an action does not affect the equilibrium risks. 
\item A larger $C_{v,\delta}$ increases the equilibrium risk, as a larger $C_{v,\delta}$ indicates that the attacker can make a larger modification on training data.
\item A larger $C_a$ decreases the equilibrium risk, as a larger $C_a$ restricts the attacker's actions to make changes.
\item A larger number of compromised nodes $| \mathcal{V}_a |$ increases the equilibrium risk as attacking more nodes gives the attacker access to modify more training samples.
\end{itemize}
} 

\label{Observation2}
\noindent
{\bf Observation 2.}{\it \ \ \ Denote the degree of node $v$ as $\mid \mathcal{B}_v \mid / (\mid \mathcal{V} \mid -1)$ and the degree of a network as the average degree of all the nodes. The impact of network topologies are summarized as follows.
\begin{itemize}
\item Networks with higher degrees and fewer nodes are less vulnerable to attacker.
\item Balanced networks, i.e., nodes in these networks have the same number of neighboring nodes, are more secure than unbalanced networks. 
\end{itemize}
Notice that here we assume that each node in the network contains the same number of samples.  
} 

\label{Observation3}
\noindent
{\bf Observation 3.}{\it \ \ \ For a specified network, assuming that all the nodes contain the same number of samples, the impact of a node is summarized as follows.
\begin{itemize}
\item Nodes with higher degrees are more vulnerable, i.e., attacking nodes with higher degrees leads to a higher global equilibrium risk. 
\item Attacking nodes with lower degrees can lead to a higher global equilibrium risk if the network contains nodes with higher degrees, comparing to networks without high degree nodes but has the same average degree.  
\end{itemize}
} 

Observations 1, 2 and 3 summarize our previous numerical experiments in \cite{Zhang}. From Observation 1, the attacker makes a larger impact when he has a higher capability, such as, he has a larger $C_{v,\delta}$ and a smaller $C_a$, or he can attack more nodes. From Observation 2, on the one hand, the attacker can choose to attack unbalanced networks with lower degrees and more nodes to make a more significant impact on the learner, on the other hand, the learner should select balanced networks with higher degrees and fewer nodes to reduce potential damages from attacker.  From Observation 3, the attacker benefits more from attacking nodes with higher degrees, while the learner should avoid using high degree nodes. These observations provide both players the strategies to make a larger impact on the other ones. In the following subsections, we present in detail how the attacker and the learner can find better strategies against each other. 

\subsection{Attacker's Strategies} 

Consider that a DSVM learner operates training data on an unbalanced network. We assume that the attacker knows the learner's algorithm and the network topology. We also assume that the attacker has the ability to attack any nodes in this network with 
$\sum_{v=1}^{V_a} C_{v,\delta} \leq C_{V_a,\delta}$, i.e., a total sum of all changed values in the network should be bounded by $C_{V_a,\delta}$. Notice that bounded $C_{V_a,\delta}$ represents a trade-off between attacking more nodes $V_a$ and attacking each nodes with larger $C_{v,\delta}$. Since attacking different nodes leads to different global equilibrium risks, and the attacker tends to higher risks, there exists an optimal strategy of selecting $\mathcal{V}_a$ and $\{ C_{v,\delta} \}_{v\in\mathcal{V}_a}$ for the attacker which has the highest equilibrium global risk with a bounded $C_{V_a,\delta}$. The optimal strategy can be found by solving the following problem:

\begin{equation}
\label{eq:OptimalAttacker}
\begin{array}{l}
\begin{array}{*{20}{l}}
{\mathop{\max} \limits_{\{\mathcal{V}_a, C_{v,\delta} \}} \mathop {\min }\limits_{\left\{ {{{\bf{w}}_v},{b_v},\{\xi_{vn}\}} \right\}} \mathop {\max }\limits_{\{ {\delta _{vn}}\} } \frac{1}{2}\sum\limits_{v \in \mathcal{V}} {{{\left\| {{{\bf{w}}_v}} \right\|_2^2}}}  + V{C_l}\sum\limits_{v \in \mathcal{V}} {\sum\limits_{n = 1}^{{N_v}} {{\xi _{vn}}} } }\\
{\begin{array}{*{20}{c}}
{}&{}
\end{array} \ \ \ \ \ \ \ \ \ \ \ \ \ \ -{C_a}\sum\limits_{v \in \mathcal{V}_a} {\sum\limits_{n = 1}^{{N_v}} {{{\left\| {{\delta _{vn}}} \right\|}_0}} } }- \sum\limits_{v \in \mathcal{V}_a} h_v
\end{array}\\
{\rm{s}}.{\rm{t}}.\\
\begin{array}{*{20}{l}}
{\begin{array}{*{20}{l}}
{{{\rm{y}}_{vn}}({\bf{w}}_v^T{{\bf{x}}_{vn}} + {b_v}) \ge 1 - {\xi _{vn}},}\\{{{\rm{y}}_{vn}}({\bf{w}}_v^T({{\bf{x}}_{vn}}-\delta_{vn} )+ {b_v}) \ge 1 - {\xi _{vn}},}\\
{{\xi _{vn}} \ge 0,}\\
{{{\bf{w}}_v} = {{\bf{w}}_u},{b_v} = {b_u},}\\
{{\delta _{vn}} \in {\mathcal{U}_{v}} ,i.e.,{\sum_{n = 1}^{N_v} {\left\| {{\delta _{vn}}} \right\|_2^2}  \le {C_{v,\delta} }},}\\{\sum_{v=1}^{V_a} C_{v,\delta} \leq C_{V_a,\delta}.}
\end{array}}&{\begin{array}{*{20}{l}}
{\forall v \in \mathcal{V}_l,n = 1,...,{N_v};}\\
{\forall v \in \mathcal{V}_a,n = 1,...,{N_v};}\\
{\forall v \in \mathcal{V},n = 1,...,{N_v};}\\
{\forall v \in \mathcal{V},u \in {\mathcal{B}_v};}\\
{\forall v \in {\mathcal{V}_a};}\\{}
\end{array}}
\end{array}
\end{array}
\end{equation}

Note that Problem (\ref{eq:OptimalAttacker}) extends Problem (\ref{eq:alminmax}) by maximizing over variables $\mathcal{V}_a$ and $\{C_{v,\delta} \}$ with a new constraint $\sum_v^{V_a} C_{v,\delta} \leq C_{V_a,\delta}$ that captures a bound of on attacker's ability. The last term $h_v $ in the objective function represents the cost of attacking node $v$. 

Problem (\ref{eq:OptimalAttacker}) is based on the assumption that the attacker has the knowledge of the learner's algorithm and the network topology. The learner aims to minimize the classification errors in Problem (\ref{eq:alminmax}), while the attacker maximizes that errors. In Problem (\ref{eq:OptimalAttacker}), the attacker has two components to maximize. Maximizing over $\{ \delta_{vn} \}$ is the same as in Problem (\ref{eq:alminmax}). Maximizing over $\mathcal{V}_a$ and $\{C_{v,\delta} \}$ indicates the objective of the attacker to maximize equilibrium risk of the original game with a bounded $C_{V_a,\delta}$ and a cost $h_v$. By solving Problem (\ref{eq:OptimalAttacker}), the attacker can find the optimal strategy of $\mathcal{V}_a$ and $\{C_{v,\delta}  \}_{v\in\mathcal{V}_a}$, which has the maximized equilibrium risk. 

\begin{figure}
\centering
\subfigure[]{
\includegraphics[width=0.3\textwidth]{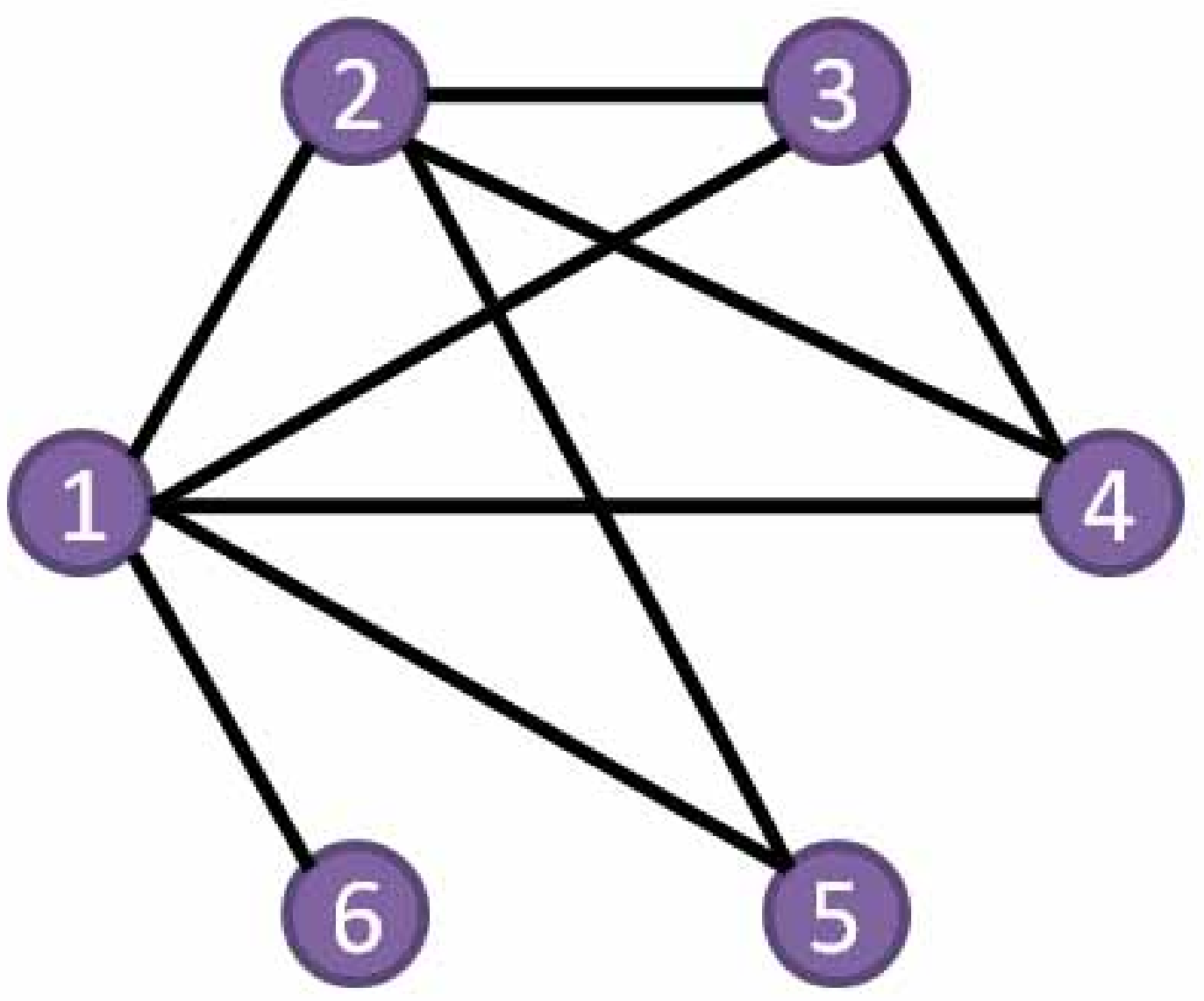}}
\subfigure[]{
\includegraphics[width=0.4\textwidth]{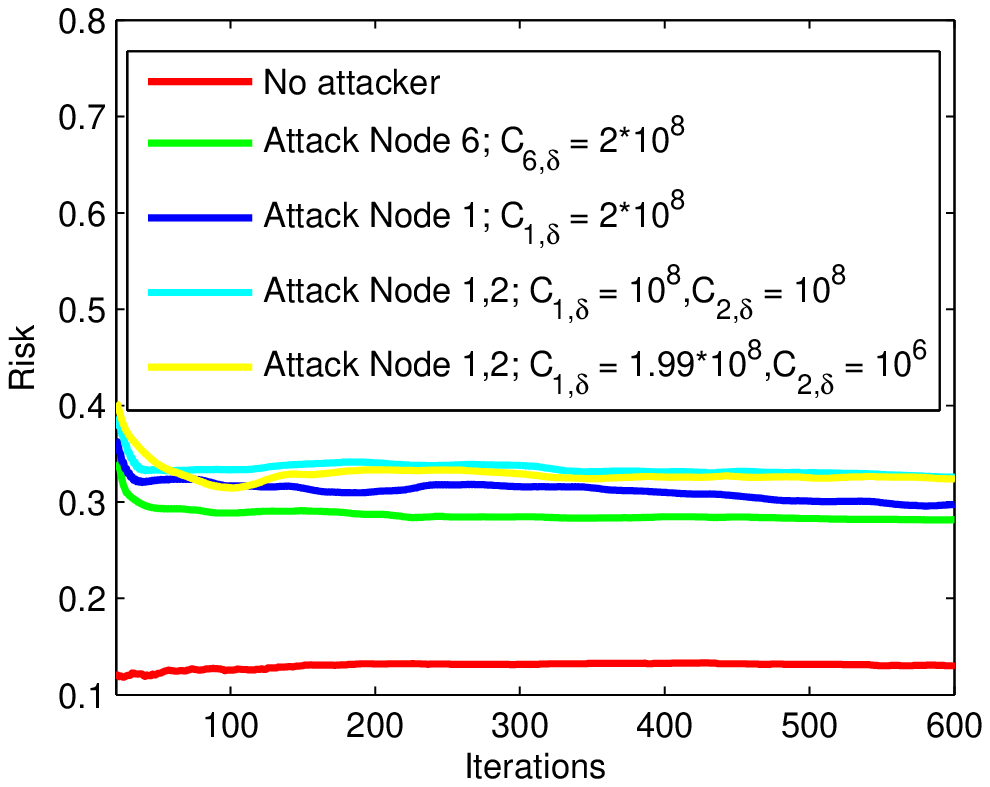}}
\vspace{-3mm}
\caption{Evolution of moving average of global empirical risks of ADMoM-DSVM with the attacker on Spam dataset \cite{Spambase}. Each node contains $40$ training samples. Attacker has four strategies with same $C_{V_a,\delta} = 2 \times 10^8$ and $C_a = 0.01$.}
\label{fig:OptimalAttacker}
\end{figure}

However, solving Problem (\ref{eq:OptimalAttacker}) can be a challenge as the decision variables $\mathcal{V}_a$ and $ C_{v,\delta}$ are coupled with the decisions of the learner and the attacker. The attacker is still able to make a larger impact on the learner by Observation 1, 2 and 3. For example, instead of randomly picking nodes to attack and assigning $C_{v,\delta}$, the attacker can strategically attack high degree nodes, which leads to a higher risk from observations. One numerical example is shown in Fig. \ref{fig:OptimalAttacker}. 

Consider the learner operates on a network shown in Fig. \ref{fig:OptimalAttacker}(a). We assume that the attacker can only attack $2$ nodes with the bound $C_{V_a,\delta} = 2\times 10^8$, and the cost of attacking node $v$, i.e., $h_v$ are the same for every node. A naive attacker may randomly attack $1$ node with $C_{v,\delta} = 2\times 10^8$. However, a smart attacker will choose $2$ nodes with higher degrees, and by modifying the value of $C_{v,\delta}$ in both nodes, he can make a larger impact on the learner. Numerical results are shown in Fig. \ref{fig:OptimalAttacker}(b).

From Fig. \ref{fig:OptimalAttacker}, the attacker has four different strategies, $(i)$ the attacker only attacks Node 6, $(ii)$ the attacker only attacks Node 1, $(iii)$ the attacker attacks Node 1,2 with balanced ability, and $(iv)$ the attacker attacks Node 1,2 with unbalanced ability. We can see that when the attacker choose Strategy $(iii)$, the risk is the highest. However, if we take the cost of attacking different nodes into consideration, this strategy may not be the best as attacking $2$ nodes may cost too much. But from the example, we can see that Observation 1, 2 and 3 provides us a way to find a better strategy for the attacker. They also provide us tools of finding better strategies for the learner.

\subsection{Learner's Strategies}

A DSVM learner aims to find the best discriminant function{s} with the least classification errors. Since an attacker will increase the classification errors, a better strategy of the learner is to reduce the attacker's impact as much as possible. In this section, we assume that the learner is trying to find the strategy of network topology that has a smallest risk with potential attacks. We assume that the learner has the ability to select any kinds of network topologies and assign any number of training samples in each node.  The learner's strategy can be found by solving the following problem, 

\begin{equation}
\label{eq:OptimalLearner}
\begin{array}{l}
\begin{array}{*{20}{l}}
{\mathop{\min} \limits_{\{\mathcal{V}, \mathcal{B}_v, N_v \}} \mathop {\min }\limits_{\left\{ {{{\bf{w}}_v},{b_v},\{\xi_{vn}\}} \right\}} \mathop {\max }\limits_{\{ {\delta _{vn}}\} } \frac{1}{2}\sum\limits_{v \in \mathcal{V}} {{{\left\| {{{\bf{w}}_v}} \right\|_2^2}}}  + V{C_l}\sum\limits_{v \in \mathcal{V}} {\sum\limits_{n = 1}^{{N_v}} {{\xi _{vn}}} } }\\
{\begin{array}{*{20}{c}}
{}&{}
\end{array} \ \ \ \ \  \ \ \ -{C_a}\sum\limits_{v \in \mathcal{V}_a} {\sum\limits_{n = 1}^{{N_v}} {{{\left\| {{\delta _{vn}}} \right\|}_0}} } }- \sum\limits_{v\in \mathcal{V}} T_v(N_v) -\sum\limits_{v\in\mathcal{V}} B_v(\mathcal{B}_v)
\end{array}\\
{\rm{s}}.{\rm{t}}.\\
\begin{array}{*{20}{l}}
{\begin{array}{*{20}{l}}
{{{\rm{y}}_{vn}}({\bf{w}}_v^T{{\bf{x}}_{vn}} + {b_v}) \ge 1 - {\xi _{vn}},}\\{{{\rm{y}}_{vn}}({\bf{w}}_v^T({{\bf{x}}_{vn}}-\delta_{vn} )+ {b_v}) \ge 1 - {\xi _{vn}},}\\
{{\xi _{vn}} \ge 0,}\\
{{{\bf{w}}_v} = {{\bf{w}}_u},{b_v} = {b_u},}\\
{{\delta _{vn}} \in {\mathcal{U}_{v}} ,i.e.,{\sum_{n = 1}^{N_v} {\left\| {{\delta _{vn}}} \right\|_2^2}  \le {C_{v,\delta} }},}
\end{array}}&{\begin{array}{*{20}{l}}
{\forall v \in \mathcal{V}_l,n = 1,...,{N_v};}\\
{\forall v \in \mathcal{V}_a,n = 1,...,{N_v};}\\
{\forall v \in \mathcal{V},n = 1,...,{N_v};}\\
{\forall v \in \mathcal{V},u \in {\mathcal{B}_v};}\\
{\forall v \in {\mathcal{V}_a}.}
\end{array}}
\end{array}
\end{array}
\end{equation}

Note that Problem (\ref{eq:OptimalLearner}) extends Problem (\ref{eq:alminmax}) by minimizing over variables $\mathcal{V},\mathcal{B}_v$ and $N_v$ with new costs $T_v(N_v)$ and $B_v(\mathcal{B}_v)$. $T_v(N_v)$ represents the cost of training $N_v$ samples in node $v$, $B_v(\mathcal{B}_v)$ represents the cost of sending information from node $v$ to his neighboring nodes $u\in\mathcal{B}_v$. Problem (\ref{eq:OptimalLearner}) can be understood as the learner's attention to minimize equilibrium risk of the game with potential attacks by finding the best network topology $\mathcal{V}$, $\mathcal{B}_v$ and training samples' assignments $N_v$. 

Solving Problem (\ref{eq:OptimalLearner}) can be a challenge as $\mathcal{V}$, $\mathcal{B}_v$ and $N_v$ are coupled with the decisions of the learner and the attacker. But the learner can benefit from Observation 1, 2 and 3. For example, the learner should select a balanced network with fewer nodes and higher degree, which has a smaller equilibrium risk. However, in reality, the learner may not be able to modify network topologies as the connections between nodes can be fixed, or it may not be possible to add connections between nodes. Thus, to reduce the impact of the attacker, the learner requires actionable defense strategies. 

In the following sections, we present four different defense strategies, and we verify their effectiveness with numerical experiments. 

\section{DSVM Defense Strategies}

In this section, we present four defense strategies (DSs) for the DSVM learner. We show their effectiveness with numerical experiments. 

\subsection{DSVM Defense Strategy 1: Selecting Network Topology}

\begin{figure}
\centering
\subfigure[]{
\includegraphics[width=0.3\textwidth]{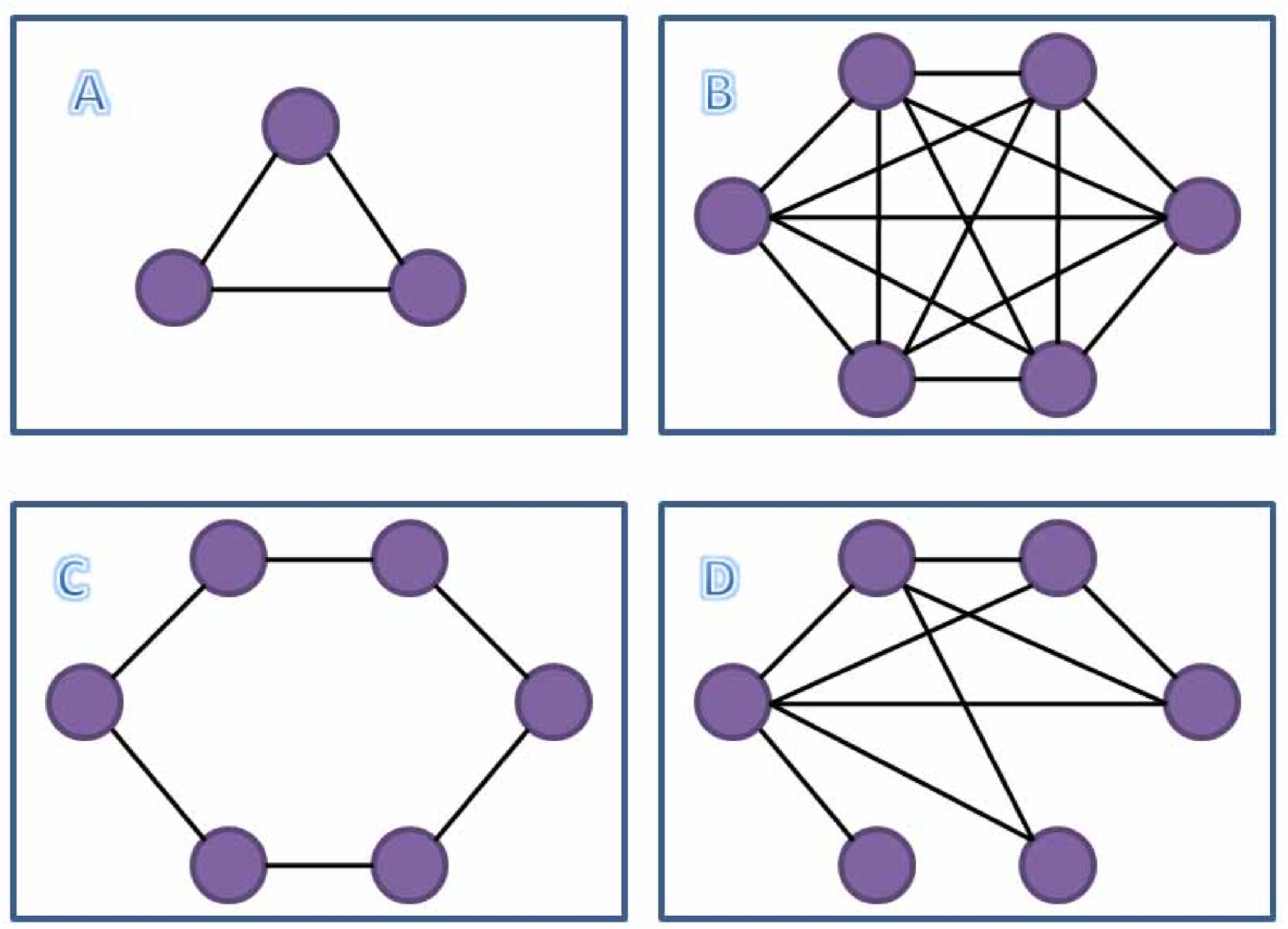}}
\subfigure[]{
\includegraphics[width=0.4\textwidth]{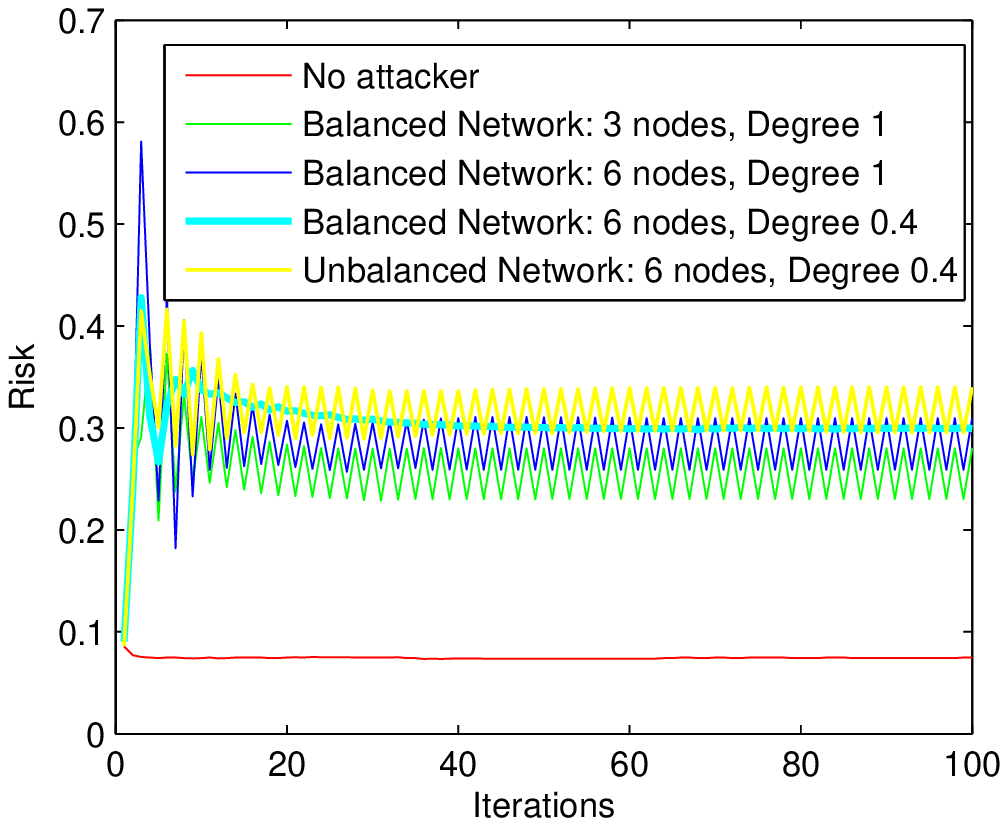}}
\vspace{-3mm}
\caption{Evolution of the global empirical risks of ADMoM-DSVM with the attacker on a random dataset. The learner has four options of network topologies which are shown in figure (a). Topology $A$ is a balanced network with $3$ nodes and degree $1$, each node in this network contains $80$ training samples. Network $B$ is a balanced network with $6$ nodes and degree $1$. Network $C$ is a balanced network with $6$ nodes and degree $0.4$. Network $D$ is an unbalanced network with $6$ nodes and degree $0.4$. Each node in network $B,C$ and $D$ contains $40$ training samples. Attacker attacks $1$ node in network $A$, but he attacks $2$ nodes in network $B,C,D$, so the attacker can modify the same number of training samples in different network topologies. The attacker has $C_{v,\delta} = 5\times 10^5$ and $C_a = 0.01$. }  
\label{fig:DefenseNetwork}
\end{figure}

DS 1 for the learner is to find a network topology that has a smaller risk when there is an attacker. From the last section, the learner can find the network topology by solving Problem (\ref{eq:OptimalLearner}). However, Problem (\ref{eq:OptimalLearner}) is difficult to solve. But we are still able to find a secure network topology using Observation 2 and 3. The network topology should be close to a balanced network with fewer nodes and a higher degree. A numerical experiment is shown in Fig. \ref{fig:DefenseNetwork}.

Consider that a DSVM learner trains 300 samples, and he aims to select a secure network topology from four topologies shown in Fig. \ref{fig:DefenseNetwork}(a). DS 1 indicates that we should select network $A$ or $B$ as network $A$ has the smallest number of nodes among all the networks, and network $B$ has the highest degree among networks $B,C,D$. Numerical results in Fig. \ref{fig:DefenseNetwork}(b) show that DS 1 has smaller risks. 

{
Though selecting a network with fewer nodes reduces the vulnerability of the learner, but each node is required to train more training samples, which takes more time and memory usages. In addition, the learner may not have the ability to select a proper network topology as most networks are given. And improving the degree of the network may not be always applicable as adding connections between nodes is costly. Thus DS 1 is suitable for cases when the network connections are convenient to modify. }

{
Consider the application in which several wireless temperature sensors in the building aim to decide whether to open their air conditioners or not. Since a large building may have hundreds of sensors and the temperatures are always changing with time, centralized classifications may take a significant amount of time to collect, transmit, and process the data. DSVMs can be used here as each sensor operates on its own data, and only a small amount of information is transmitted between sensors. But if there is an attacker who has the ability to modify the training data in several sensors, then the sensors in the building will lead to wrong decisions. In this case, wireless temperature sensors can adapt and modify their network topology. Thus, a secure strategy here is to use DS 1 to create a balanced network with fewer sensors and a higher average degree.}

\subsection{DSVM Defense Strategy 2: Adding Training Samples} 

\begin{figure}
\centering
\subfigure[Defense starts from step $0$.]{
\includegraphics[width=0.4\textwidth]{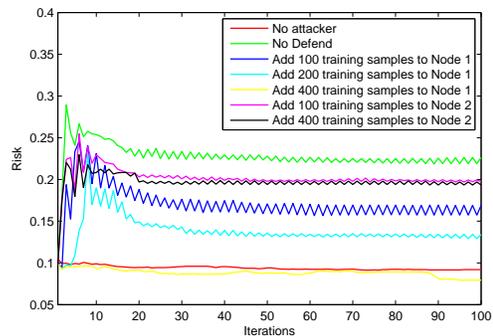}}
\subfigure[Defense starts from step $50$.]{
\includegraphics[width=0.4\textwidth]{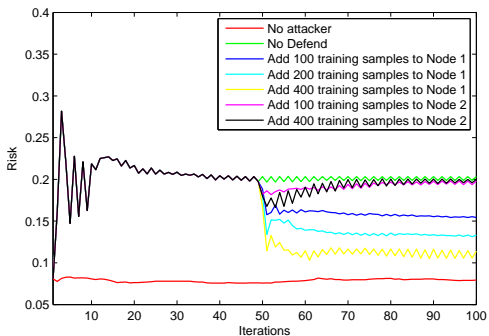}}
\vspace{-3mm}
\caption{Evolution of global empirical risks of ADMoM-DSVM with an attacker at a balanced network with $6$ nodes and degree $0.4$ on random dataset, which is shown in Fig. \ref{fig:DefenseNetwork}(a) as Network $C$. Each node contains $40$ training samples.  Attacker only attacks node $1$ with $C_{1,\delta} = 10^6$ and $C_a = 0.01$.}
\label{fig:DefenseAdd}
\end{figure}

Since the attacker is limited to making modifications on the training data, a higher volume of training data will decrease the ratio of incorrect data at a node. As long as most of the data are correct, the learner can find the discriminant function with small classification errors. Thus adding more training samples becomes a reasonable defense strategy. Numerical experiments are shown in Fig. \ref{fig:DefenseAdd}. 

From Fig. \ref{fig:DefenseAdd}, when we add training samples to network, the risk is lower. Thus adding training samples is a proper defense strategy. Note that more samples we add, the lower the risk will be. Adding training samples to compromised nodes turns out to be more efficient than adding to uncompromised nodes. However, training more samples requires more time and memory usages, which sacrifices efficiency. Thus, DS 2 is a trade-off between efficiency and security. 

{
DS 2 is suitable for the case when the learner cannot change the network topology, but the size of training data is sufficiently large and each node has a strong computing capability. For example, consider an application where several environmental stations plan to detect whether some areas are under pollution with a wired communication network. DSVMs are suitable to process a large amount of data computations and transmissions. However, if an attacker modifies the training data, environmental stations may lead to misdetection. In this case, DS 1 may not be applicable as the wired connections between each station are fixed. However, since each station can collect enough training data and has a higher computation capability, DS 2 is more appropriate and the learner can add more training samples to each node to make the training process more secure. Note that using more samples requires additional time to train and more space to store data.}

\subsection{DSVM Defense Strategy 3: Verification Method}

DS 1 suggests that the learner uses a balanced network with fewer nodes and a higher degree. However, using fewer nodes requires that each node trains more training samples, which sacrifices the efficiency. Increasing the degree of the network requires creating more connections between nodes, which are usually not applicable as building new lines may incur a high cost. DS 2 indicates that adding more training samples can reduce the vulnerability of the network, which also sacrifices the efficiency. Thus, both DS 1 and DS 2 have their limitations on securing a training process. In this section, we present a verification method that reduces the vulnerability without modifying the network topology or adding training samples. 

In ADMoM-DSVM Algorithm $1$, each node in the network receives $\mathbf{r}_u$ from his neighboring nodes and it also sends his $\mathbf{r}_v$ to his neighboring nodes at each step. Since $\mathbf{r}_u$ from neighboring nodes of node $v$ contributes to the updates of $\mathbf{r}_v$, a wrong $\mathbf{r}_u$ can lead to an incorrect update of $\mathbf{r}_v$. As a result, if node $v$ is protected from receiving wrong $\mathbf{r}_u$ from compromised nodes, it can have a correct discriminant function. 

Recall DSVM Problem (\ref{eq:DSVMMatrix}), note that consensus constraints $\mathbf{r}_v = \omega_{vu}, \omega_{vu}=\mathbf{r}_u$ force all the local decision variables $\mathbf{r}_v$ to agree with each other. Thus, $\mathbf{r}_1^{(t)}\approx...\approx\mathbf{r}_V^{(t)}$ should hold for every step $t$ during the training process. Thus, if $\mathbf{r}_v$ violates this, then the learner can tell that node $v$ is under attack. With Algorithm $1$, if node $v$ finds $\mathbf{r}_u$ is significantly different from $\mathbf{r}_v$, then he will reject using $\mathbf{r}_u$ to update himself. We call this method as the verification method. The ADMoM-DSVM algorithm with verification method can be summarized as Algorithm $3$. 

\begin{table}
\renewcommand{\arraystretch}{1.3}
\label{tablealgorithm3}
\centering
\begin{tabular}{l}
\hline
\bfseries Algorithm 3: DSVM with Verification \\
\hline

Randomly initialize $\mathbf{r}_v^{(0)},\lambda_v^{(0)},\omega_{vu}^{(0)}$, set $\alpha_v^{(0)}=\mathbf{0}_{(p+1)\times 1}$,\\ set $\widehat{\mathcal{B}_v} = \mathcal{B}_v$.\\   
1:\ \ \bf{for} $t=0,1,2,...$ do\\
2:\ \ \ \ \ \ \ \ \bf{for all} $v\in \mathcal{V}$ do \\
3:\ \ \ \ \ \ \ \ \ \ \ \ \ Compute $\lambda_v^{(t+1)}$ via (\ref{eq:DSVMSoli1}) with $\widehat{\mathcal{B}_v}$.\\
4:\ \ \ \ \ \ \ \ \ \ \ \ \ Compute $\mathbf{r}_v^{(t+1)}$ via (\ref{eq:DSVMSoli2}) with $\widehat{\mathcal{B}_v}$.\\
5:\ \ \ \ \ \ \ \ \bf{end for} \\
6:\ \ \ \ \ \ \ \ \bf{for all} $v\in \mathcal{V}$ do \\
7:\ \ \ \ \ \ \ \ \ \ \ \ Broadcast $\mathbf{r}_v^{(t+1)}$ to all neighbors $u\in \mathcal{B}_v$.\\
8:\ \ \ \ \ \ \ \bf{end for}\\
9:\ \ \ \ \ \ \ \bf{for all} $v\in \mathcal{V}$ do \\
10:\ \ \ \ \ \ \ \ \ \ \ \ Set $\widehat{\mathcal{B}}_v = \emptyset$. \\
11:\ \ \ \ \ \ \ \ \ \ \ \ \bf{for all} $u\in \mathcal{B}_v$ do \\
12:\ \ \ \ \ \ \ \ \ \ \ \ \ \ \ \ \ \bf{if} $ \mid 1 - \frac{{\parallel\mathbf{r}_u^{(t+1)}\parallel}_2}{{\parallel\mathbf{r}_v^{(t+1)}\parallel}_2} )  \mid < \tau $\\
13:\ \ \ \ \ \ \ \ \ \ \ \ \ \ \ \ \ \ \ \ \ \ Set $u\in\widehat{\mathcal{B}_v}$. \\
14:\ \ \ \ \ \ \ \ \ \ \ \ \ \ \ \ \ \bf{end if}\\
15:\ \ \ \ \ \ \ \ \ \ \ \ \bf{end for}\\
16:\ \ \ \ \ \ \ \bf{end for}\\
17:\ \ \ \ \ \ \ \bf{for all} $v\in \mathcal{V}$ do \\
18:\ \ \ \ \ \ \ \ \ \ \ \ \ Compute $\omega_{vu}^{(t+1)}$ via (\ref{eq:DSVMSoli3}) with $\widehat{\mathcal{B}_v}$.\\
19:\ \ \ \ \ \ \ \ \ \ \ \ \ Compute $\alpha_v^{(t+1)}$ via (\ref{eq:DSVMSoli4}) with $\widehat{\mathcal{B}_v}$.\\
20:\ \ \ \ \ \ \ \bf{end for}\\
21:\ \bf{end for}\\
\hline
\end{tabular}
\end{table}  

\begin{figure}
\centering
\subfigure[Attack starts from step $0$]{
\includegraphics[width=0.4\textwidth]{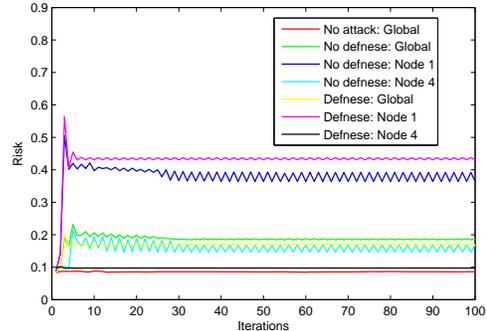}}
\subfigure[Attack starts from step $60$]{
\includegraphics[width=0.4\textwidth]{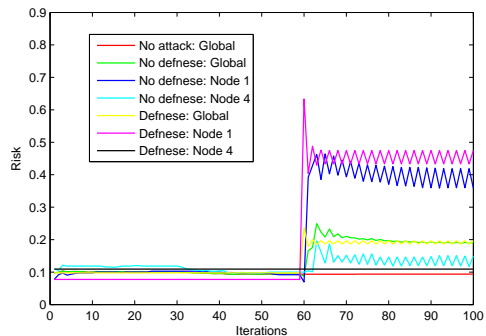}}
\vspace{-3mm}
\caption{Evolution of the empirical risks of ADMoM-DSVM with the attacker at a balanced network with $4$ nodes degree $0.4$ on random dataset. Each node contains $60$ training samples. Attacker only attacks Node 1 with $C_{1,\delta} = 10^5$ and $C_a = 0.01$. $\tau = 0.1$.}
\label{fig:DefenseTrust1}
\end{figure}

\begin{figure}
\centering
\subfigure[Attack starts from step $0$]{
\includegraphics[width=0.4\textwidth]{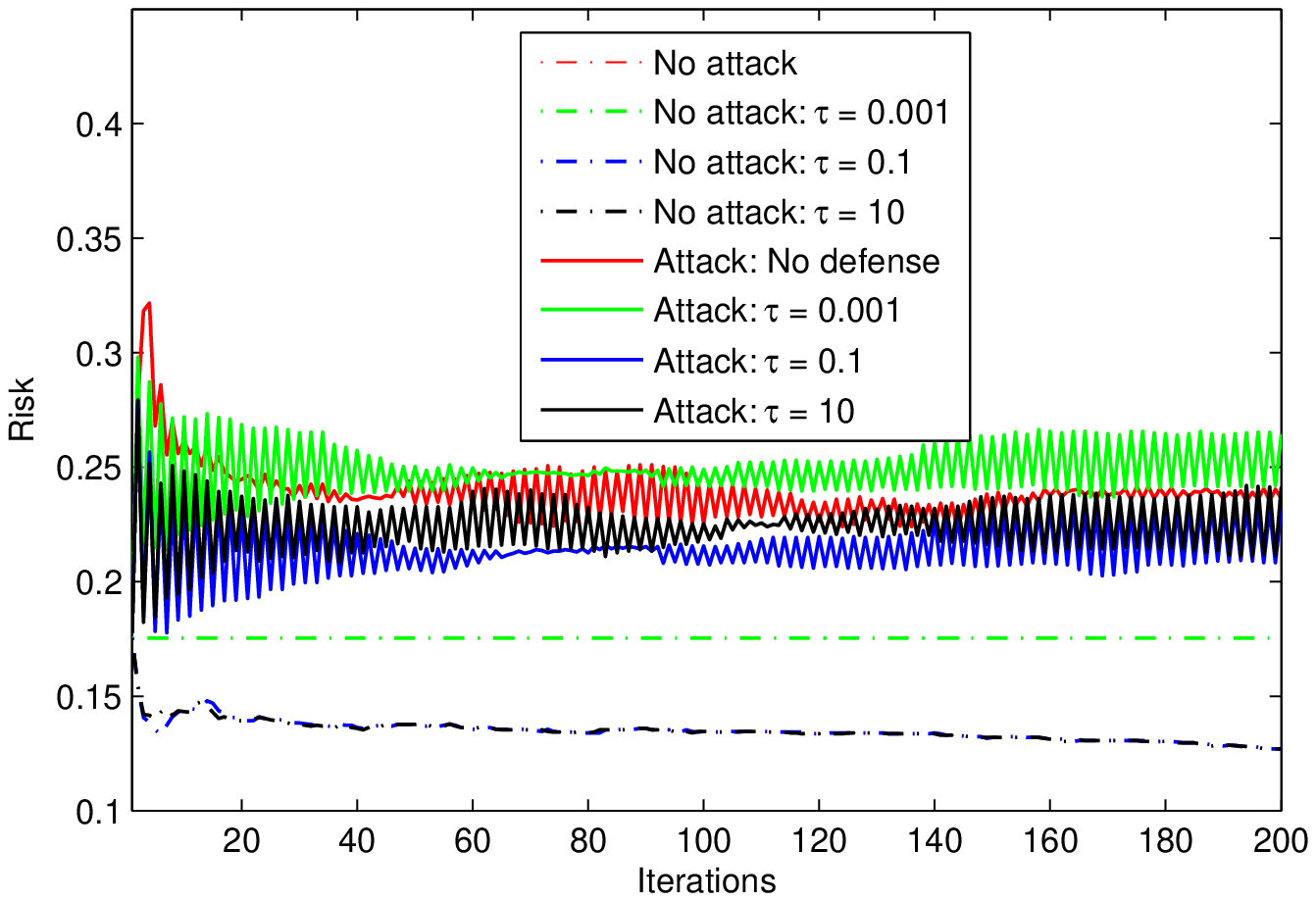}}
\subfigure[Attack starts from step $120$]{
\includegraphics[width=0.4\textwidth]{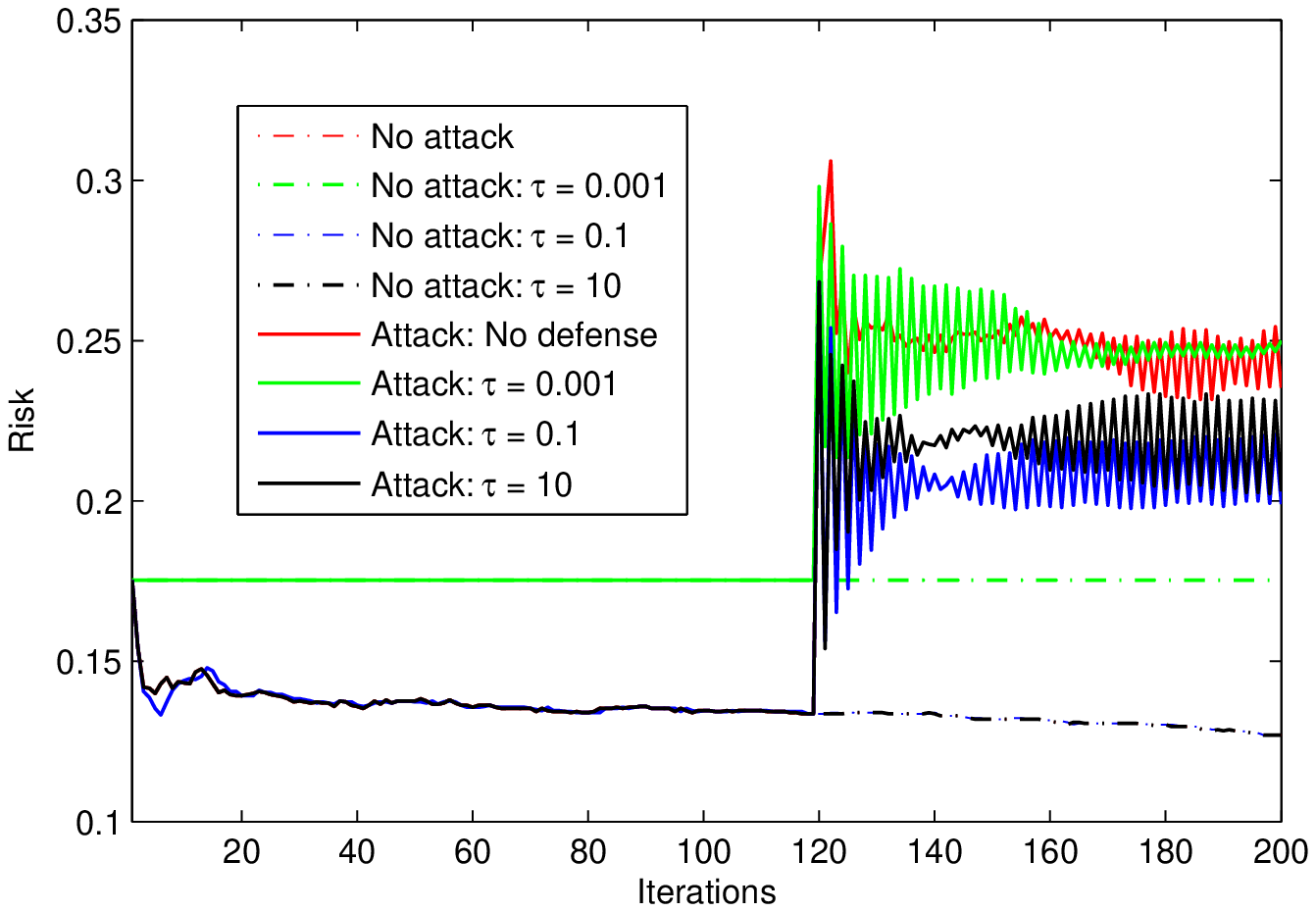}}
\vspace{-3mm}
\caption{Evolution of the global empirical risks of ADMoM-DSVM with the attacker at a balanced network with $4$ nodes degree $0.4$ on Spambase dataset \cite{Spambase}. Each node contains $60$ training samples. Attacker only attacks node $1$ with $C_{1,\delta} = 10^6$ and $C_a = 0.01$ . }
\label{fig:DefenseTrust2}
\end{figure}

Algorithm $3$ differs from Algorithm $1$ in the verification method. Each node computes with information only from trusted neighboring nodes $u\in\widehat{\mathcal{B}}_v$. The verification method is based on the inequality in step $12$ of Algorithm 3. $\tau$ indicates the tolerance of indifference from $\mathbf{r}_u$ to $\mathbf{r}_v$, and $\tau \geq 0$. When $\tau$ is close to $0$, node $v$ is very sensitive to the information from other nodes, and it only uses $\mathbf{r}_u$ that is very close to $\mathbf{r}_v$. Numerical experiments are shown in Fig. \ref{fig:DefenseTrust1} and Fig. \ref{fig:DefenseTrust2}.

We can see from  Fig. \ref{fig:DefenseTrust1} that the global risk has decreased when there is a verification method. Note that in uncompromised node $4$, the risk is closed to the risk when there is no attacker, while in compromised node $1$, the risk is higher than the risk when there is no defense. This indicates that, though the verification method protects uncompromised nodes from receiving misleading information, it also prevents compromised nodes from receiving correct information. 

Fig. \ref{fig:DefenseTrust2} compares the global risks when the learner uses different $\tau$. We can see that when $\tau = 10$, the risk is higher than the risk when $\tau = 0.1$, thus some of the misleading information is still able to be spread in the network. When $\tau = 0.001$, we can see that the risk is even higher than the risk when there is no defense. Also note that when there is no attacker, the risk of DSVM with $\tau = 0.001$ does not converge to the risk of normal DSVM. This indicates that when $\tau$ is close to $0$, the misleading information cannot be spread to other nodes, but the useful information is also forbidden to transmit. Thus DS 3 requires a proper selection of $\tau$. 

{
DS $3$ is suitable for the case when training data are used in a large network. Since it is difficult for the attacker to attack many nodes at the same time, for a network with a large number of nodes, all the uncompromised nodes can be kept from being affected by the compromised nodes. Moreover, the learner can distinguish compromised nodes by their high local classification risks, and thus, without revoking the training process and retraining all the data in every node, the learner is able to maintain the resilience of the training process by deleting or correcting the compromised nodes. Comparing to DS $1$ and $2$, DS 3 does not sacrifice efficiency to maintain security, but the compromised nodes may result in worse performances.}

\subsection{DSVM Defense Strategy 4: Rejection Method}

DSs $1$, $2$ and $3$ have shown that with selecting proper network topologies, adding training samples and verification method, DSVM learner can be less vulnerable to attacks. However, DSs $1$ and $2$ will sacrifice efficiency. In DS $3$, compromised nodes may result in worse performances. In this section, we present the rejection method where each node rejects unreasonable updates. With the rejection method, once there is an attacker, the iteration will terminate to prevent further damages caused by the attacker.   

The rejection method relies on a combined residual, which measures both the primal and dual error simultaneously:

\begin{equation}
\label{eq:Residual}
{J^{(t+1)}} = \eta \sum\limits_{v \in {\cal V}} {\sum\limits_{u \in {B_v}} {\left\| {  \omega _{vu}^{(t+1)} - \omega _{vu}^{(t)} } \right\|_2^2} }  + \frac{2}{\eta }\sum\limits_{v \in {\cal V}} {\left\| { \alpha _v^{(t+1)} -  \alpha _v^{(t)}} \right\|_2^2}.
\end{equation}

Note that the combined residual contains two terms. The first term measures the dual residual. The second term measures the primal residual. The combined residual has the following lemma \cite{He}.

\label{Lemma4}
\noindent
{\bf Lemma 4.}{\it \ \ \ Iterations (\ref{eq:DSVMSoli1})-(\ref{eq:DSVMSoli4}) satisfy that $J^{(t+1)}\le J^{(t)}$, which can also be rewritten as: 
\begin{equation}
\label{eq:Converge}
\begin{array}{l}
\eta \sum\limits_{v \in \mathcal{V}} {\sum\limits_{u \in {B_v}} {\left\| {\omega _{vu}^{(t + 1)} - \omega _{vu}^{(t)}} \right\|_2^2} }  + \frac{2}{\eta }\sum\limits_{v \in \mathcal{V}} {\left\| {\alpha _v^{(t + 1)} - \alpha _v^{(t)}} \right\|_2^2} \\
 \le \eta \sum\limits_{v \in\mathcal{V}} {\sum\limits_{u \in {B_v}} {\left\| {\omega _{vu}^{(t)} - \omega _{vu}^{(t - 1)}} \right\|_2^2} }  + \frac{2}{\eta }\sum\limits_{v \in \mathcal{V}} {\left\| {\alpha _v^{(t)} - \alpha _v^{(t - 1)}} \right\|_2^2}.
\end{array}
\end{equation}
} 
 
A proof of Lemma $4$ can be found in \cite{He}. Lemma 4 indicates that the combined residual always decreases. Since the attacker aims to break the training process, this inequality will not be satisfied when there is an attacker. Note that computing Inequality (\ref{eq:Converge}) requires $\omega_{vu}$ and $\alpha_v$ from every node, which can be down by a fusion center in centralized machine learning problems. However, since the learner uses a fully distributed network without a fusion center, we decentralize Inequality (\ref{eq:Converge}) into $|\mathcal{V}|$ distributed inequalities, for $v \in \mathcal{V}$:

\begin{equation}
\label{eq:Convergev}
\begin{array}{l}
\eta {\sum\limits_{u \in {B_v}} {\left\| {\omega _{vu}^{(t + 1)} - \omega _{vu}^{(t)}} \right\|_2^2} }  + \frac{2}{\eta } {\left\| {\alpha _v^{(t + 1)} - \alpha _v^{(t)}} \right\|_2^2} \\
 \le \eta {\sum\limits_{u \in {B_v}} {\left\| {\omega _{vu}^{(t)} - \omega _{vu}^{(t - 1)}} \right\|_2^2} }  + \frac{2}{\eta } {\left\| {\alpha _v^{(t)} - \alpha _v^{(t - 1)}} \right\|_2^2}.
\end{array}
\end{equation}

Note that there is no guarantee that Inequality (\ref{eq:Convergev}) holds based on Inequality (\ref{eq:Converge}). As a result, we relax the distributed inequality with a parameter $\rho >1$, which is summarized in the following proposition. 

\label{Proposition1}
\noindent
{\bf Proposition 1.}{\it \ \ \ Iterations (\ref{eq:DSVMSoli1})-(\ref{eq:DSVMSoli4}) satisfy that $J_v^{(t+1)}\le \rho J_v^{(t)}$, where 
\begin{equation}
\label{eq:ConvergevRelax}
\begin{array}{l}
J_v^{(t)} = \eta  {\sum\limits_{u \in {B_v}} {\left\| {\omega _{vu}^{(t)} - \omega _{vu}^{(t - 1)}} \right\|_2^2} }  + \frac{2}{\eta } {\left\| {\alpha _v^{(t)} - \alpha _v^{(t - 1)}} \right\|_2^2}.
\end{array}
\end{equation}
} 
 
\begin{proof}

Let us assume that $J_v^{(t+1)}\le \rho J_v^{(t)}$ does not hold for $v = v_0$, we have $J_{v_0}^{(t+1)}> \rho J_{v_0}^{(t)}$ and $J_{v \neq v_0}^{(t+1)} \leq \rho J_{v\neq v_0}^{(t)}$. As a result,  $J_{v_0}^{(t+1)}> \rho^{(t+1)} J_{v_0}^{(0)}$ which increases exponentially with $\rho>1$. Since $J_v^{(t)}$ is always larger than $0$, 
Inequality (\ref{eq:Converge}) will be violated eventually. Proposition 1 holds. 
\end{proof}

\begin{figure}
\centering
\subfigure[Attack starts from step $0$]{
\includegraphics[width=0.4\textwidth]{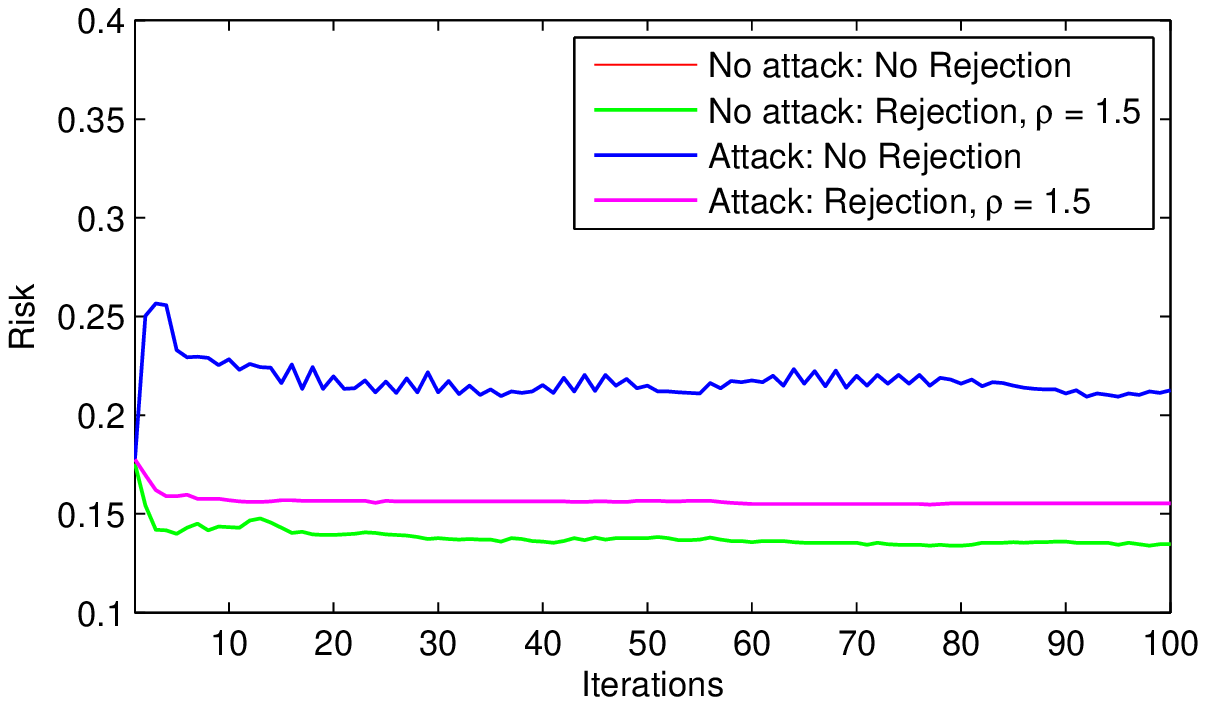}}
\subfigure[Attack starts from step $60$]{
\includegraphics[width=0.4\textwidth]{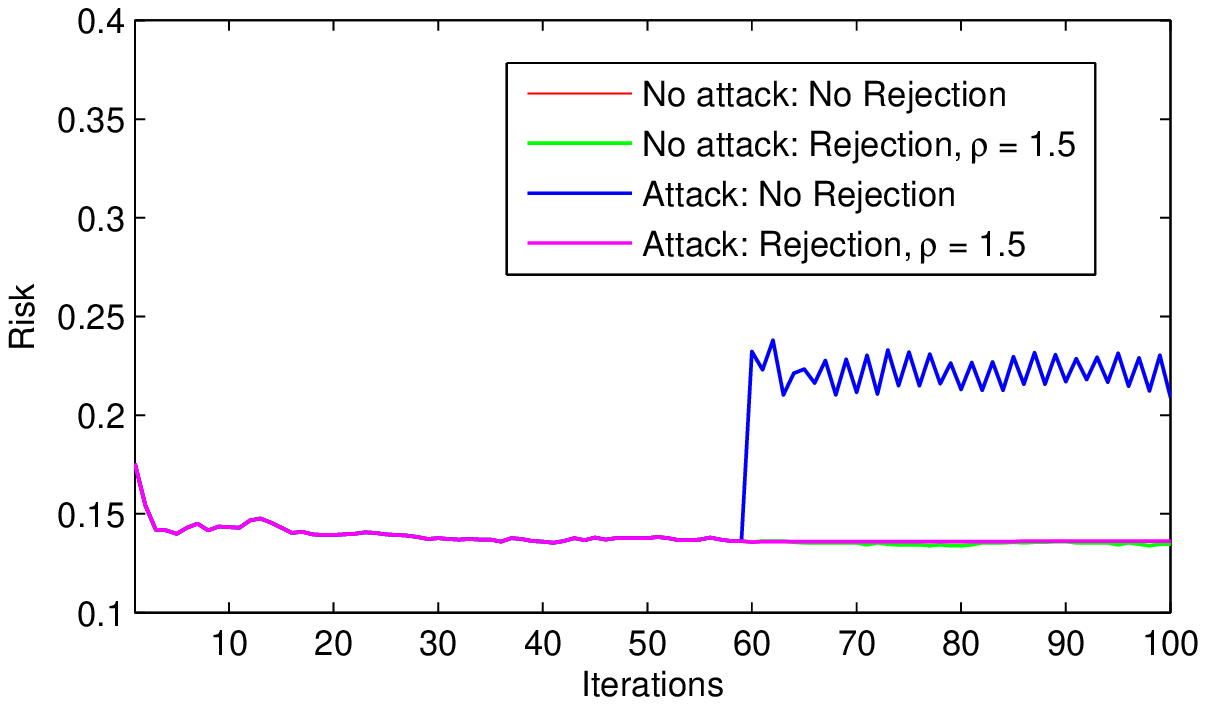}}
\vspace{-3mm}
\caption{Evolution of the empirical risks of ADMoM-DSVM Rejection with the attacker at a balanced network with $4$ nodes of degree $0.4$ on Spambase dataset \cite{Spambase}. Each node has $60$ training samples. The attacker only attacks 1 node with $C_{1,\delta} = 10^5$ and $C_a = 0.01$. The rejection method has $\rho = 1.5$. }
\label{fig:DefenseCG15}
\end{figure}

\begin{figure}
\centering
\subfigure[Attack starts from step $0$]{
\includegraphics[width=0.4\textwidth]{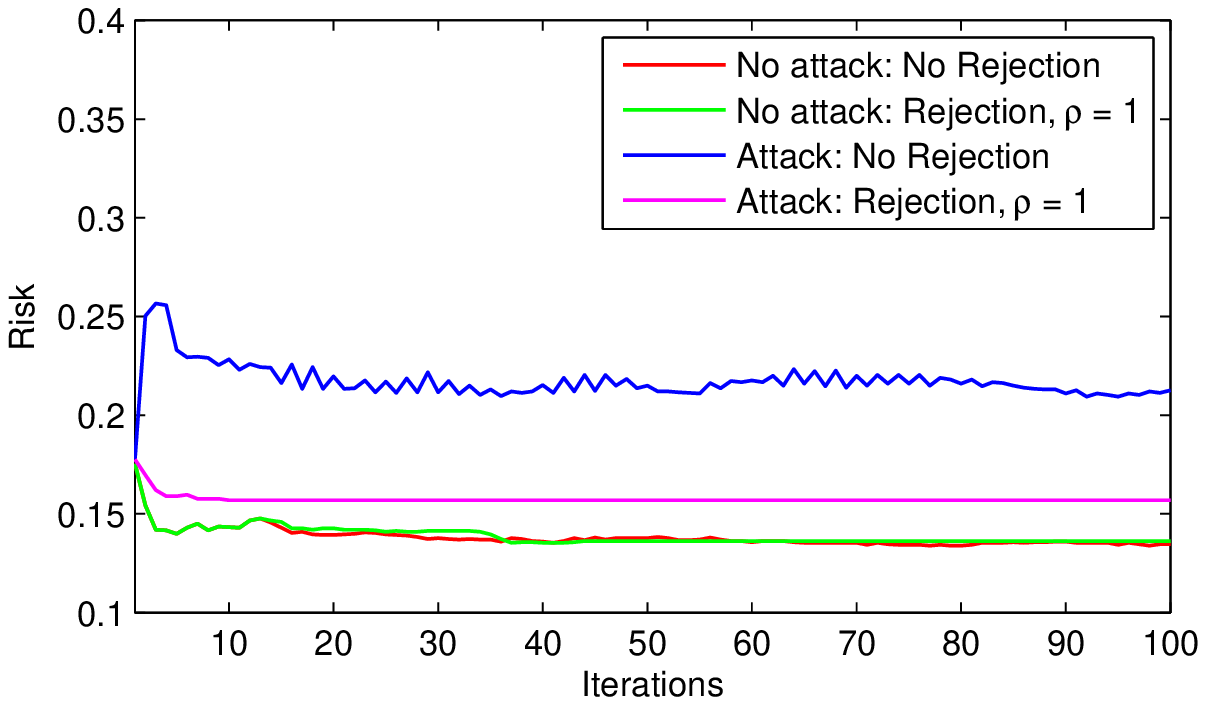}}
\subfigure[Attack starts from step $60$]{
\includegraphics[width=0.4\textwidth]{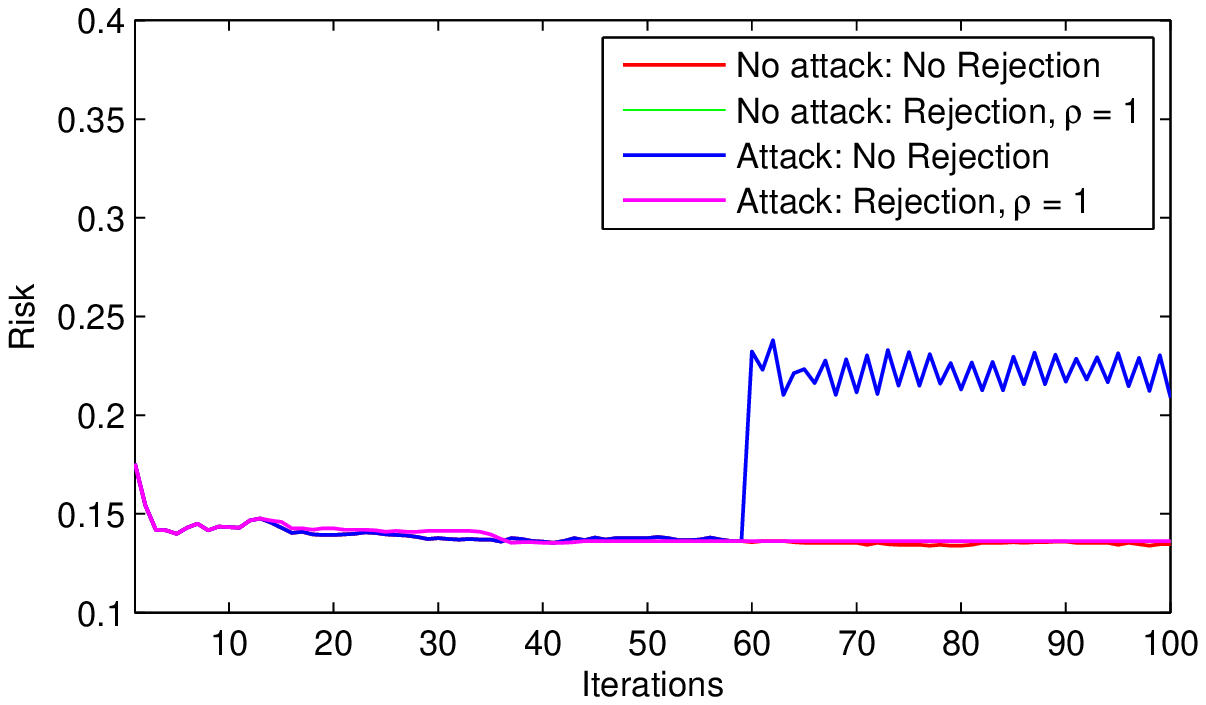}}
\vspace{-3mm}
\caption{Evolution of the empirical risks of ADMoM-DSVM Rejection with the attacker at a balanced network with $4$ nodes of degree $0.4$ on Spambase dataset \cite{Spambase}. Each node has $60$ training samples. The attacker only attacks 1 node with $C_{1,\delta} = 10^5$ and $C_a = 0.01$. The rejection method has $\rho = 1$. }
\label{fig:DefenseCG1}
\end{figure}

\begin{figure}
\centering
\subfigure[Attack starts from step $0$]{
\includegraphics[width=0.4\textwidth]{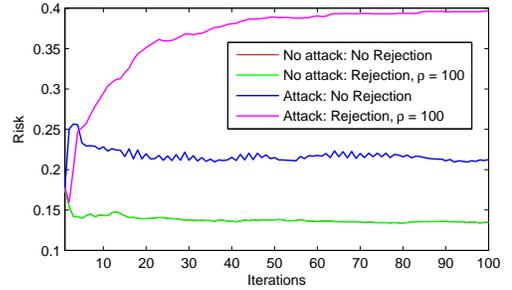}}
\subfigure[Attack starts from step $60$]{
\includegraphics[width=0.4\textwidth]{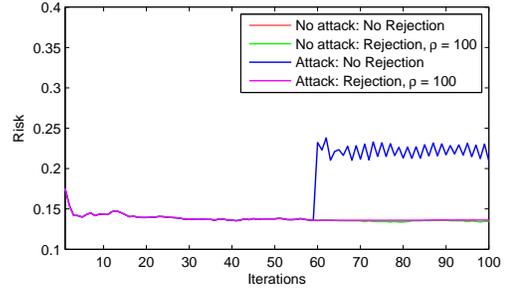}}
\vspace{-3mm}
\caption{Evolution of the empirical risks of ADMoM-DSVM Rejection with the attacker at a balanced network with $4$ nodes of degree $0.4$ on Spambase dataset \cite{Spambase}. Each node has $60$ training samples. The attacker only attacks 1 node with $C_{1,\delta} = 10^5$ and $C_a = 0.01$. The rejection method has $\rho = 100$. }
\label{fig:DefenseCG100}
\end{figure}

With the inequality in Proposition 1, the new DSVM algorithm with rejection method can be summarized into Algorithm $4$. In Algorithm $4$, if the inequality at Step 15 is not satisfied, then the update will be rejected. $J_v^{(0)}$ should be set to be sufficiently large to pass the first rejection test. Numerical experiments are shown in Fig. \ref{fig:DefenseCG15}, Fig. \ref{fig:DefenseCG1}, and Fig. \ref{fig:DefenseCG100}.

\begin{table}
\renewcommand{\arraystretch}{1.3}
\label{tablealgorithm4}
\centering
\begin{tabular}{l}
\hline
\bfseries Algorithm 4: DSVM with Rejection \\
\hline

Randomly initialize $\mathbf{r}_v^{(0)},\lambda_v^{(0)},\omega_{vu}^{(0)}$ and $\alpha_v^{(0)}=\mathbf{0}_{(p+1)\times 1}$,\\ set $J_v^{(0)}$ very large.\\
1:\ \ \bf{for} $t=0,1,2,...$ do\\
2:\ \ \ \ \ \ \ \ \bf{for all} $v\in \mathcal{V}$ do \\
3:\ \ \ \ \ \ \ \ \ \ \ \ \ Compute $ \lambda_v^{(t+1)}$ via (\ref{eq:DSVMSoli1}).\\
4:\ \ \ \ \ \ \ \ \ \ \ \ \ Compute $\mathbf{r}_v^{(t+1)}$ via (\ref{eq:DSVMSoli2}).\\
5:\ \ \ \ \ \ \ \ \bf{end for} \\
6:\ \ \ \ \ \ \ \ \bf{for all} $v\in \mathcal{V}$ do \\
7:\ \ \ \ \ \ \ \ \ \ \ \ Broadcast $\mathbf{r}_v^{(t+1)}$ to all neighbors $u\in \mathcal{B}_v$.\\
8:\ \ \ \ \ \ \ \bf{end for}\\
9:\ \ \ \ \ \ \ \bf{for all} $v\in \mathcal{V}$ do \\
10:\ \ \ \ \ \ \ \ \ \ \ \ \ Compute $\omega_{vu}^{(t+1)}$ via (\ref{eq:DSVMSoli3}).\\
11:\ \ \ \ \ \ \ \ \ \ \ \ \ Compute $\alpha_{v}^{(t+1)}$ via (\ref{eq:DSVMSoli4}).\\
12:\ \ \ \ \ \ \ \ \ \ \ \ \ Compute $J_{v}^{(t+1)}$ via (\ref{eq:ConvergevRelax}).\\
13:\ \ \ \ \ \ \ \bf{end for}\\
14:\ \ \ \ \ \ \ \bf{for all} $v\in \mathcal{V}$ do \\
15:\ \ \ \ \ \ \ \ \ \ \ \ \bf{if} $J_v^{(t+1)} > \rho J_v^{(t)}$ \\
16:\ \ \ \ \ \ \ \ \ \ \ \ \ \ \ \ \ \ $\lambda_v^{(t+1)} = \lambda_v^{(t)}$, $\mathbf{r}_v^{(t+1)} = \mathbf{r}_v^{(t)}$, \\
17:\ \ \ \ \ \ \ \ \ \ \ \ \ \ \ \ \ \ $\alpha_v^{(t+1)} = \alpha_v^{(t)}$, $\omega_{vu}^{(t+1)} = \omega_v^{(t)}$,\\
18:\ \ \ \ \ \ \ \ \ \ \ \ \ \ \ \ \ \ $J_v^{(t+1)} = J_v^{(t)}$.\\
19:\ \ \ \ \ \ \ \ \ \ \ \ \bf{end if}\\
20:\ \ \ \ \ \ \ \bf{end for} \\
21:\ \bf{end for}\\
\hline
\end{tabular}
\end{table}  


From Fig. \ref{fig:DefenseCG15}, we can see that the DSVM algorithm with rejection method has a lower risk than the normal algorithm when there is an attacker. And it has the same performance when there is no attacker, which indicates that when $\rho =1.5$, rejection method does not affect the training process. Fig. \ref{fig:DefenseCG1} and Fig. \ref{fig:DefenseCG100} show the results when $\rho = 1$ and $\rho = 100$, respectively. We can see from Fig. \ref{fig:DefenseCG1} that when $\rho = 1$, the risk is lower when there is an attacker, but convergence slows down when there is no attacker. We can see from Fig. \ref{fig:DefenseCG100} that when $\rho = 100$, the risk with rejection method is even higher than the risk of the standard algorithm, because wrong updates can still be treated as a correct update and accumulates as iteration goes. 

From the numerical experiments, the value of $\rho$ is important to the rejection method. A smaller $\rho$ may slow down the convergence of the DSVM algorithm without attacker, a larger $\rho$ does not prevent attacks. With a properly selected $\rho$, the training process becomes less vulnerable to attackers.

{
DS $4$ is suitable for a wide range of applications as wrong updates will be rejected. Comparing to DSs $1$ and $2$, DS $4$ does not sacrifice efficiency. Comparing to DS $3$, compromised nodes in DS $4$ has been kept from being further damaged by the attacker. One possible drawback of DS $4$ is that it may require insights of the problem to find a proper $\rho$. }

{
Each defense strategy is suitable for a different scenario and applications. The choice of defense strategies will depend on the applications and the constraints on the defender's actions. Though four defense strategies have their own advantages and disadvantages, a combination of all the defense strategies can be used to secure the training process of the learner.   }

\section{Conclusion}

Distributed support vector machines are ubiquitous but inherently vulnerable to adversaries. This paper has investigated defense strategies of DSVM against potential attackers. We have established a game-theoretic framework to capture the strategic interactions between an attacker and a learner with a network of distributed nodes. We have shown that the nonzero-sum game is strategically equivalent to a zero-sum game, and hence, its equilibrium can be characterized by a saddle-point equilibrium solution to a minimax problem. By using the technique of ADMoM, we have developed secure and resilient algorithms that can respond to the adversarial environment. We have shown that a balanced network with fewer nodes and a higher degree is less vulnerable to the attacker. Moreover, adding more training samples has been proved to reduce the vulnerability of the system. We have shown that verification method where each node verifies information from neighboring nodes can protect uncompromised nodes from receiving misleading information, but compromised nodes are also prevented from receiving correct information. We have shown that rejection method where each node rejects unreasonable updates can stop global training process from deterioration, thus wrong information is thwarted from affecting the system. One direction of future works is to extend the current framework to investigate nonlinear DSVM and other machine learning algorithms.

{
\section*{Appendix A: Proof of Lemma 2}
A detailed proof of Lemma 2 can be found in our previous work \cite{Zhang}. By using hinge loss function, we reformulate Problem (10) into the following problem:
\begin{equation}
\label{eq:alminmax1}
\begin{array}{l}
\begin{array}{*{20}{l}}
{\mathop {\min }\limits_{\left\{ {{{\bf{w}}_v},{b_v}} \right\}} \mathop {\max }\limits_{\left\{ {{\delta _{vn}}} \right\}}  \frac{1}{2}\sum\limits_{v \in \mathcal{V}} {{{\left\| {{{\bf{w}}_v}} \right\|_2^2}}} }\\
{\begin{array}{*{20}{c}}
{}&{}
\end{array} + {V_l}{C_l}\sum\limits_{v \in \mathcal{V}_l} {\sum\limits_{n = 1}^{{N_v}} {{{\left[ {1 - {{\rm{y}}_{vn}}({\bf{w}}_v^T{{\bf{x}}_{vn}} + {b_v})} \right]}_ + }} } }\\
{\begin{array}{*{20}{c}}
{}&{}
\end{array} + {V_a}{C_l}\sum\limits_{v \in \mathcal{V}_a} {\sum\limits_{n = 1}^{{N_v}} {{{\left[ {1 - {{\rm{y}}_{vn}}({\bf{w}}_v^T({{\bf{x}}_{vn}} - {\delta _{vn}}) + {b_v})} \right]}_ + }} } }\\
{\begin{array}{*{20}{c}}
{}&{}
\end{array} - {C_a}\sum\limits_{v \in \mathcal{V}_a} {\sum\limits_{n = 1}^{{N_v}} {{{\left\| {{\delta _{vn}}} \right\|}_0}} } }
\end{array}\\
{\rm{s}}{\rm{.t}}{\rm{.   }}\begin{array}{*{20}{c}}
{\begin{array}{*{20}{l}}
{{{\bf{w}}_v} = {{\bf{w}}_u},{b_v} = {b_u},}\\
{({\delta _{v1}},...,{\delta _{v{N_v}}}) \in {{\cal U}_v},}
\end{array}}&{\begin{array}{*{20}{l}}
{\forall v \in {\cal V},u \in {\mathcal{B}_v};}\\
{\forall v \in {{\cal V}_a}.}
\end{array}}
\end{array}
\end{array}
\end{equation}
Similarly, Problem (\ref{eq:MinMax}) can be reformulated into the following problem:
\begin{equation}
\label{eq:MinMaxHinge}
\begin{array}{*{20}{c}}
\begin{array}{l}
\mathop {\min }\limits_{\left\{ {{{\bf{w}}_v},{b_v}} \right\}} \mathop {\max }\limits_{\{ {\delta _v}\} } \frac{1}{2}\sum\limits_{v \in \mathcal{V}} {{{\left\| {{{\bf{w}}_v}} \right\|}^2}} \\
\begin{array}{*{20}{c}}
{}&{}&{}
\end{array} + {V_l}{C_l}\sum\limits_{v \in\mathcal{V}_l} {\sum\limits_{n = 1}^{{N_v}} {{{\left[ {1 - {{\rm{y}}_{vn}}({\bf{w}}_v^T{{\bf{x}}_{vn}} + {b_v})} \right]}_ + }} } \\
\begin{array}{*{20}{c}}
{}&{}&{}
\end{array} + {V_a}{C_l}\sum\limits_{v \in \mathcal{V}_a} {\sum\limits_{n = 1}^{{N_v}} {{{\left[ {1 - {{\rm{y}}_{vn}}({\bf{w}}_v^T{{\bf{x}}_{vn}} + {b_v})} \right]}_ + }} } \\
\begin{array}{*{20}{c}}
{}&{}&{}
\end{array} + \sum\limits_{v \in\mathcal{V}_a}^{} {\left( {{V_a}{C_l}{\bf{w}}_v^T{\delta _v} - {C_a}\left\| {{\delta _v}} \right\|_0} \right)} 
\end{array}\\
{\begin{array}{*{20}{l}}
{}\\
{{\rm{s}}.{\rm{t}}.\begin{array}{*{20}{c}}
{\begin{array}{*{20}{c}}
{{{\bf{w}}_v} = {{\bf{w}}_u},{b_v} = {b_u},}&{\forall v \in {\cal V},u\in\mathcal{B}_v;}\\
{{{\delta _v} \in {\mathcal{U}_{v0}},}}&{\forall v \in {{\cal V}_a}.}
\end{array}}
\end{array}}
\end{array}}
\end{array}
\end{equation}
As a result, we only need to prove that problem (\ref{eq:alminmax1}) is equivalent to problem (\ref{eq:MinMaxHinge}). Since both of problems are min-max problems with the same variables, we only need to prove that we minimize the same maximization problem. Moreover, Moreover, since $\{\delta_{vn} \}$ is independent in the maximization part of (\ref{eq:alminmax1}), and $\delta_v$ is independent in the maximization part of (\ref{eq:MinMaxHinge}), we can separate maximization problem into $V_a$ sub-maximization problems, and solving the sub-problems is equivalent to solving the global maximization problem. As a result, we only need to show that the following sub-problem  
\begin{equation}
\label{eq:SubMaxA}
\begin{array}{l}
\mathop {\max }\limits_{\{\delta_{vn}\}\in \mathcal{U}_v} S( \{ \delta_{vn} \} ) \buildrel \Delta \over =  {V_a}{C_l}\sum\limits_{n = 1}^{{N_v}} {{{\left[ {1 - {{\rm{y}}_{vn}}({\bf{w}}_v^T({{\bf{x}}_{vn}} - {\delta _{vn}}) + {b_v})} \right]}_ + }} \\
\begin{array}{*{20}{c}}
{}&{}
\end{array} \ \ \ \ \ \ \ \ \ \ \ \ \ - {C_a}\sum\limits_{n = 1}^{{N_v}} {\left\| {{\delta _{vn}}} \right\|_0} 
\end{array}
\end{equation}
is equivalent to the following sub-problem 
\begin{equation}
\label{eq:SubMaxB}
\begin{array}{l}
\mathop {\max }\limits_{{\delta _v} \in {\mathcal{U}_{v0}}} {V_a}{C_l}\sum\limits_{n = 1}^{{N_v}} {{{\left[ {1 - {{\rm{y}}_{vn}}({\bf{w}}_v^T{{\bf{x}}_{vn}} + {b_v})} \right]}_ + }} \\
\begin{array}{*{20}{c}}
{}&{}
\end{array} + {V_a}{C_l}{\bf{w}}_v^T{\delta _v} - {C_a}\left\| {{\delta _v}} \right\|_0.
\end{array}
\end{equation}
We adopt the similar proof in \cite{Xu}, recall the properties of sublinear aggregated action set, $\mathcal{U}_v^-\subseteq\mathcal{U}_v\subseteq\mathcal{U}_v^+ $, where 
\[ \begin{array}{*{20}{c}}
{\begin{array}{*{20}{c}}
{{{\cal U}^ - } \buildrel \Delta \over = \mathop  \cup \limits_{t = 1}^n {\cal U}_t^ - ,{\rm{  }}{\cal U}_t^ -  \buildrel \Delta \over = \left\{ {\left( {{\delta _1},...,{\delta _n}} \right)\left|  \begin{array}{l}
{\delta _t} \in {{\cal U}_0};\\
{\delta _{i}} = {\bf{0}}, i \ne t.
\end{array} \right.} \right\}};\\ \medskip
{{{\cal U}^ + } \buildrel \Delta \over = \left\{ {\left( {{\alpha _1}{\delta _1},...,{\alpha _n}{\delta _n}} \right)\left| \begin{array}{l}
\sum\limits_{i = 1}^n {{\alpha _i} = 1} ;{\alpha _i} \ge 0,\\
{\delta _i} \in {{\cal U}_0},i = 1,...,n
\end{array} \right.} \right\}}.
\end{array}}
\end{array} \]
Hence, fixing any $({\bf{w}}_v,b_v)\in \mathbb{R}^{p+1}$, we have the following inequalities:
\begin{equation}
\label{eq:MainInequal}
\mathop {\max }\limits_{\{ \delta_{vn}\} \in {\cal U}_v^ - } S(\{\delta_{vn}\})  \le \mathop {\max }\limits_{\{ \delta_{vn}\} \in {\cal U}_v } S(\{\delta_{vn}\})  \le \mathop {\max }\limits_{\{ \delta_{vn}\} \in {\cal U}_v^ +  } S(\{\delta_{vn}\}) 
\end{equation}
We can show that (\ref{eq:SubMaxB}) is no larger than the leftmost term and no smaller than the rightmost term \cite{Zhang}. Thus, the equivalence between (\ref{eq:SubMaxA}) and (\ref{eq:SubMaxB}) holds. Hence, Lemma 2 holds.}

{
\section*{Appendix B: Proof of Lemma 3}
We use best response dynamics to construct the best response for the min-problem and max-problem separately. The min-problem and max-problem are achieved by fixing $\{\mathbf{r}_v,\xi_v\}$ and $\{\delta_{v}\}$, respectively.
For fixed $\{ {\bf{r}}_v^*, \xi_v^*\}$,
\begin{equation}
\label{equation8}
\begin{array}{l}
\delta _v^* \in \arg \mathop {\max }\limits_{\{\delta _v\}} \sum\limits_{v\in\mathcal{V}_a} {\left( {{V_a}{C_l}{\bf{r}_v^*}^T({{\bf{I}}_{p + 1}} - {\Pi _{p + 1}}){\delta _v}}  -C_a \| \delta_v \|_0  \right)} \\
\begin{array}{*{20}{c}}
{{\rm{s}}{\rm{.t}}{\rm{.}}}&{{\delta _v} \in {{\cal U}_{v0}}},&{\forall v \in {\cal V}_a}.
\end{array}
\end{array}
\end{equation}
We relax $l_0$ norm to $l_1$ norm to represent the cost function of the attacker. By writing the dual form of the $l_1$ norm, we arrive at 
\begin{equation}
\label{equation2545}
\begin{array}{l}
\delta _v^* \in \arg \mathop {\max }\limits_{\left\{ {{\delta _v},{s_v}} \right\}} {V_a}{C_l}{\bf{r}_v^*}^T({{\bf{I}}_{p + 1}} - {\Pi _{p + 1}}){\delta _v} - {{\bf{1}}^T}{s_v}\\
{\rm{s}}.{\rm{t}}.{\rm{   }}\begin{array}{*{20}{c}}
{\begin{array}{*{20}{c}}
{{C_a}{\delta _v} \le {s_v},}\\
{{C_a}{\delta _v} \ge  - {s_v},}\\
{{\delta _v} \in {{\cal U}_{v0}}.}
\end{array}}
\end{array}
\end{array}
\end{equation}
For fixed $\{ \delta_v^* \}$, we have 
\begin{equation}
\label{equation13}
\begin{array}{*{20}{c}}
{\begin{array}{*{20}{l}}
{\mathop {\min }\limits_{\left\{ {{{\bf{r}}_v},{\omega _{vu}},{{\bf{\xi }}_v}} \right\}} \frac{1}{2}\sum\limits_{v \in \mathcal{V}}{{\bf{r}}_v^T({{\bf{I}}_{p + 1}} - {\Pi _{p + 1}}){{\bf{r}}_v}} }\\
{\begin{array}{*{20}{c}}
{}&{}
\end{array} + {V_a}{C_l}\sum\limits_{v\in\mathcal{V}_a   } {{\bf{r}}_v^T({{\bf{I}}_{p + 1}} - {\Pi _{p + 1}})\delta _v^*}  + V{C_l}\sum\limits_{v\in\mathcal{V}} {{{\bf{1}}_v^T{{\bf{\xi }}_v}} } }
\end{array}}\\
{}\\
{{\rm{s}}.{\rm{t}}.\begin{array}{*{20}{c}}
{\begin{array}{*{20}{c}}
{{{\bf{Y}}_v}{{\bf{X}}_v}{{\bf{r}}_v} \ge {{\bf{1}}_v} - {{\bf{\xi }}_v}},&{\forall v \in \mathcal{V};}\\
{{{\bf{\xi }}_v} \ge {{\bf{0}}_v}},&{\forall v \in \mathcal{V};}\\
{{{\bf{r}}_v} = {\omega _{vu}},{\omega _{vu}} = {{\bf{r}}_u}},&{\forall v \in \mathcal{V},\forall u \in {\mathcal{B}_u}.}
\end{array}}
\end{array}}
\end{array}
\end{equation}
Note that term ${ - {C_a }\left\| {\delta _v^*} \right\|_0}$ is removed since it does not play a role in the minimization problem. Based on (\ref{equation2545}) and (\ref{equation13}), we have the method of solving Problem (\ref{eq:MinMaxMatrix}) as follows, first step is that we randomly pick initial $\{ \mathbf{r}_v^{(0)}, \delta_{v}^{(0)} \}$, and then we solve Max-problem (\ref{equation2545}) with $\{ \mathbf{r}_v^{(0)} \}$ to obtain $\{\delta_{v}^{(1)}\}$. In next step, we solve Min-problem (\ref{equation13}) to obtain $\{ \mathbf{r}_v^{(1)} \}$ with $\{\delta_{v}^{(1)}\}$ from the previous step. We repeat solving the max-problem with $\{ \mathbf{r}_v^{(t-1)} \}$ and solving the min-problem with $\{\delta_{v}^{(t)}\}$ until convergence. Furthermore, we use the alternating direction method of multipliers (ADMoM) to solve Problem (\ref{equation13}).}

{
The ADMoM is a distributed optimization algorithm solving the following problem:
\begin{equation}
\label{equationB1(37)}
\begin{array}{l}
\mathop {\min }\limits_{{\bf{r}},\omega} {f}({\bf{r}}) + {g}(\omega )\\
\begin{array}{*{20}{c}}
{{\rm{s}}{\rm{.t}}{\rm{.}}}&{\begin{array}{*{20}{c}}
{{\bf{Mr}} = \omega, }
\end{array}}
\end{array}
\end{array}
\end{equation}
where $f$ and $g$ are convex functions \cite{Eckstein}.}

{
The augmented Lagrangian corresponding to (\ref{equationB1(37)}) is 
\begin{equation}
\label{equationB2(38)}
L({\bf{r}},\omega ,\alpha ) = {f}({\bf{r}}) + {g}(\omega ) + {\alpha ^T}({\bf{Mr}} - \omega ) + \frac{\eta }{2}{\left\| {{\bf{Mr}} - \omega } \right\|^2},
\end{equation}
where $\alpha$ denotes the Lagrange multiplier.}

{
Then, the ADMoM solves problem (\ref{equationB1(37)}) by the update rules below:
\begin{equation}
\label{equationB3(39)}
{\bf{r}}^{(t + 1)} \in \arg \mathop {\min }\limits_{{\bf{r}} } L({\bf{r}},\omega^{(t)},\alpha^{ (t)});
\end{equation}
\begin{equation}
\label{equationB4(40)}
\omega^{ (t + 1)} \in \arg \mathop {\min }\limits_{\omega } L({\bf{r}}^{(t + 1)},\omega ,\alpha^{ (t)});
\end{equation}
\begin{equation}
\label{equationB5(41)}
\alpha ^{ (t + 1)} = \alpha ^{(t)} + \eta ({\bf{Mr}}^{(t + 1)} - \omega ^{ (t + 1)}).
\end{equation}
The objective here is to transform Problem (\ref{equation13}) into the form of (\ref{equationB1(37)}), and then we can solve Problem (\ref{equation13}) by iterations (\ref{equationB3(39)}), (\ref{equationB4(40)}), and (\ref{equationB5(41)}). We adopt a similar method in \cite{Forero}, which leads to the following result.}

\noindent{
{\bf Remark 1.}{\it \ \ Each node iterates $\lambda_v^{(t)},\mathbf{r}_v^{(t)}$ and $\alpha_v^{(t)}$, given by
\begin{equation}
\label{eq:MinMaxSoli2In}
\begin{array}{*{20}{l}}
{{\lambda _v^{(t+1)}}}
{ \in \arg \mathop {\max }\limits_{{\bf{0}} \le {{\bf{\lambda }}_v} \le VC_l{{\bf{1}}_v}}  - \frac{1}{2}\lambda _v^T{{\bf{Y}}_v}{{\bf{X}}_v}{\bf{U}}_v^{ - 1}{\bf{X}}_v^T{{\bf{Y}}_v}{\lambda _v}}\\
{\begin{array}{*{20}{c}}
{}&{}
\end{array} \ \ \ \ \ \ \ \ \  \ \ \ \ \ + {{({{\bf{1}}_v} + {{\bf{Y}}_v}{{\bf{X}}_v}{\bf{U}}_v^{ - 1}{{\bf{f}}_v^{(t)}})}^T}{\lambda _v}},
\end{array}
\end{equation}
\begin{equation}
\label{eq:MinMaxSoli3In}
{{{\bf{r}}_v^{(t+1)}} = {\bf{U}}_v^{ - 1}\left( {{\bf{X}}_v^T{{\bf{Y}}_v}{\lambda _v^{(t+1)}} - {{\bf{f}}_v^{(t)}}} \right)},
\end{equation}
\begin{equation}
\label{eq:MinMaxSoli4In}
\omega_{vu}^{(t+1)} = \frac{1}{2} (\mathbf{r}_v^{(t+1)} + \mathbf{r}_u^{(t+1)}),
\end{equation}
\begin{equation}
\label{eq:MinMaxSoli5In}
{{\alpha _v^{(t+1)}} = {\alpha _v^{(t)}} + \frac{\eta }{2}\sum\limits_{u \in {\mathcal{B}_v}} {\left[ {{{\bf{r}}_v^{(t+1)}} - {{\bf{r}}_u^{(t+1)}}} \right]} },
\end{equation}
where $\mathbf{U}_v=(\mathbf{I}_{p+1}-\Pi_{p+1})+2\eta\vert \mathcal{B}_v\vert\mathbf{I}_{p+1},\mathbf{f}_v^{(t)}=V_a C_l\delta_v^*+2\alpha_v^{(t)}-2\eta\sum_{u\in \mathcal{U}_v}\omega_{vu}^{(t)}$.}
By combining the above remark with Problem (\ref{equation2545}), we can obtain Lemma 3.}

\bibliographystyle{ieeetr}
\bibliography{Defense.bib}

\end{document}